%% file: caring.tex
\documentclass{article}

\usepackage{microtype}
\usepackage{graphicx}
\usepackage{subfigure}
\usepackage{booktabs} 

\usepackage{hyperref}

\usepackage[accepted]{icml2024}

\usepackage{amsmath}
\usepackage{amssymb}
\usepackage{mathtools}
\usepackage{amsthm}

\usepackage[capitalize,noabbrev]{cleveref}

\input{math_commands.tex}

\usepackage{url}
\usepackage[utf8]{inputenc} 
\usepackage[T1]{fontenc}    
\usepackage{xcolor}         
\definecolor{mydarkblue}{rgb}{0,0.08,0.45}
\usepackage{hyperref}       
\hypersetup{ %
    pdftitle={},
    pdfauthor={},
    pdfsubject={},
    pdfkeywords={},
    pdfborder=0 0 0,
    pdfpagemode=UseNone,
    colorlinks=true,
    linkcolor=black,
    citecolor=mydarkblue,
    filecolor=mydarkblue,
    urlcolor=mydarkblue,
    pdfview=FitH}
\usepackage{url}            
\usepackage{booktabs}       
\usepackage{amsfonts}       
\usepackage{nicefrac}       
\usepackage{microtype}      

\usepackage{amsthm}
\usepackage{multirow}
\usepackage{bigstrut}
\usepackage{graphicx}
\usepackage{amsmath}
\usepackage{amssymb}
\usepackage{tabularx}
\usepackage{tabu}
\usepackage{siunitx}
\usepackage{adjustbox}
\usepackage{siunitx}
\usepackage{makecell}
\usepackage{colortbl}
\usepackage{color}
\usepackage{multirow}
\usepackage{blindtext}
\usepackage{comment}
\usepackage{bm}
\usepackage{bbm}
\usepackage{booktabs}
\usepackage{wrapfig}
\usepackage{array} 
\definecolor{Gray}{gray}{0.95}
\definecolor{Cyan}{rgb}{0.88,1,1}
\definecolor{LightCyan}{rgb}{0.92,1,1}
\definecolor{DarkCyan}{rgb}{0.82,1,1}
\newcommand{\ourmeos}{\textbf{\texttt{CaRiNG}} }
\newcommand{\caring}{\textbf{\texttt{CaRiNG}}}

\usepackage{tikz}
\usetikzlibrary{bayesnet}
\usetikzlibrary{arrows}
\usepackage{caption}
\usetikzlibrary{backgrounds}

\usepackage[capitalize]{cleveref}
\crefname{section}{Sec.}{Secs.}
\Crefname{section}{Section}{Sections}
\Crefname{table}{Table}{Tables}
\crefname{table}{Tab.}{Tabs.}
\newtheorem{corollary}{Corollary}
\newtheorem{definition}{Definition}
\newtheorem{lemma}{Lemma}
\newtheorem{proposition}{Proposition}
\newtheorem{theorem}{Theorem}
\usepackage{enumitem}

\usepackage{titletoc}

\newcommand{\bfsection}[1]{\noindent\textbf{#1.}}

\newcommand{\ourtitle}{\caring: Learning Temporal Causal Representation under \\ Non-Invertible Generation Process}

\usepackage[textsize=tiny]{todonotes}

\icmltitlerunning{\caring: Learning Temporal Causal Representation 
under Non-Invertible Generation Process}

\begin{document}

\twocolumn[
\icmltitle{\ourtitle}




\icmlsetsymbol{equal}{*}

\begin{icmlauthorlist}
\icmlauthor{Guangyi Chen}{equal,mbz,cmu}
\icmlauthor{Yifan Shen}{equal,mbz}
\icmlauthor{Zhenhao Chen}{equal,mbz}
\icmlauthor{Xiangchen Song}{cmu} \\
\icmlauthor{Yuewen Sun}{mbz}
\icmlauthor{Weiran Yao}{salesforce}
\icmlauthor{Xiao Liu}{mbz}
\icmlauthor{Kun Zhang}{mbz,cmu}
\end{icmlauthorlist}

\icmlaffiliation{mbz}{Mohamed bin Zayed University of Artificial Intelligence, Abu Dhabi, UAE}
\icmlaffiliation{salesforce}{Salesforce, San Francisco, US}
\icmlaffiliation{cmu}{Carnegie Mellon University, Pittsburg, US}

\icmlcorrespondingauthor{Kun Zhang}{kunz1@cmu.edu}

\icmlkeywords{Machine Learning, ICML}

\vskip 0.3in
]

\printAffiliationsAndNotice{\icmlEqualContribution}


\begin{abstract}

Identifying the underlying time-delayed latent causal processes in sequential data is vital for grasping temporal dynamics and making downstream reasoning. While some recent methods can robustly identify these latent causal variables, they rely on strict assumptions about the invertible generation process from latent variables to observed data. However, these assumptions are often hard to satisfy in real-world applications containing information loss.
For instance, the visual perception process translates a 3D space into 2D images, or the phenomenon of persistence of vision incorporates historical data into current perceptions. To address this challenge, we establish an identifiability theory that allows for the recovery of independent latent components even when they come from a nonlinear and non-invertible mix. Using this theory as a foundation, we propose a principled approach, \caring, to learn the \underline{\textbf{Ca}}usal \underline{\textbf{R}}epresentat\underline{\textbf{i}}on of \underline{\textbf{N}}on-invertible \underline{\textbf{G}}enerative temporal data with identifiability guarantees. Specifically, we utilize temporal context to recover lost latent information and apply the conditions in our theory to guide the training process. Through experiments conducted on synthetic datasets, we validate that our \ourmeos method reliably identifies the causal process, even when the generation process is non-invertible. Moreover, we demonstrate that our approach considerably improves temporal understanding and reasoning in practical applications. Code can be accessed through \url{https://github.com/sanshuiii/CaRiNG}.

\end{abstract}

\section{Introduction}

Sequential data, including video, stock, and climate observations, are integral to our daily lives. 
Gaining an understanding of the causal dynamics in such time series data has always been a crucial challenge~\citep{berzuini2012causality,ghysels2016testing,friston2009causal} and has attracted considerable attention. The core of this task is to identify the underlying causal dynamics in the data we observe.

{ Towards this goal, we focus on Independent Component Analysis (ICA)~\citep{hyvarinen2000independent}, which is a classical method for decomposing the latent signals from mixed observation. Recent advancements in nonlinear ICA~\citep{hyvarinen2016unsupervised,hyvarinen2017nonlinear,hyvarinen2019nonlinear,khemakhem2020variational,sorrenson2020disentanglement,halva2020hidden} have yielded robust theoretical evidence for the identifiability of latent variables, and enabled the use of deep neural networks to address complex scenarios. For example, by assuming the latent variables in the data generation process are mutually independent, and leveraging the auxiliary side information such as time index, domain index, or class label, \cite{hyvarinen2017nonlinear,hyvarinen2019nonlinear,halva2020hidden} have demonstrated the strong identifiability results. 
\cite{halva2021disentangling,klindt2020towards,yao2021learning,yao2022temporally,lachapelle2022disentanglement} further extend this nonlinear ICA framework into scenarios of the time-delayed dynamical systems, which allows the temporal transitions among the latent variables.  
}

\begin{figure*}[t]
\begin{center}
\centerline{\includegraphics[width=0.95\linewidth]{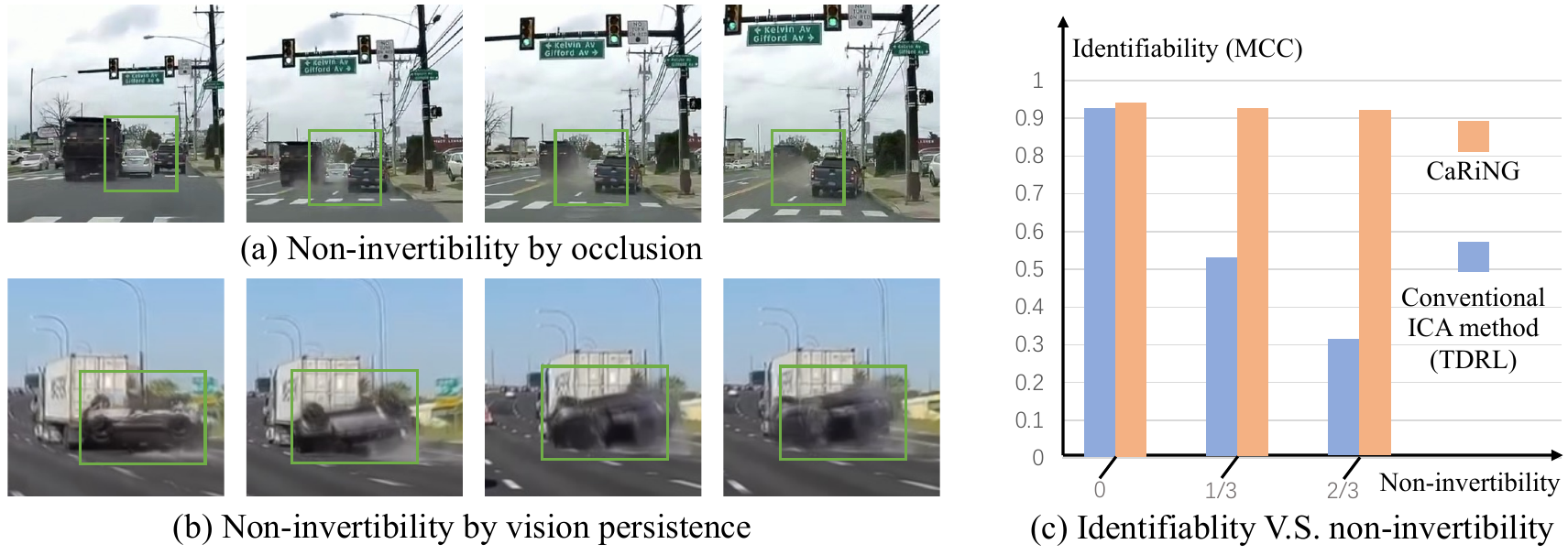}}
\vspace{-0.1in}
\caption{\textbf{Motivations of the non-invertible generation process.} (a) The occlusions raise the non-invertibility since the measured observation cannot cover the obstructed objects. (b) The vision persistence, shown with the high-speed movement of a crashing car, describes the generation process that jointly involves the current state and previous, and causes the non-invertibility. (c) The identifiability of conventional methods, such as TDRL~\cite{yao2022temporally} (blue), drops drastically with the increase of non-invertibility, while the identifiability of our method (marked in orange) still holds. 
The levels of non-invertibility are defined by removing $0, 1/3$, and $2/3$ dimensions of $\bf{z}_t$ when generating $\bf{x}_t$. For example, when the dimension of $\bf{z}_t$ is 6,  $2/3$ non-invertibility means that we remove 4 variables of $\bf{z}_t$ and use only 2 variables to generate $\bf{x}_t$.
}

\label{fig:top}
\end{center}
\vspace{-0.3in}
\end{figure*}

{However, these nonlinear ICA-based methods usually assume that the mixing function (the generation process from sources to observations) is invertible, which may be difficult to satisfy in real-world scenarios, such as the 3D to 2D projection in the visual process.
As shown in Figure~\ref{fig:top} (a) and (b), we provide two intuitive instances of the real videos to illustrate how the non-invertibility happens. 
In (a), when object occlusions occur, information from the obstructed object is lost in the generation process of the current time step, which causes non-invertibility. In (b), the persistence of vision introduces non-invertibility, since the mixing process of the current time step utilizes the history information.  
We further found that the violation of this {invertibility} assumption may cause the nonlinear ICA method to yield poor identification performance.
In part (c) of Figure~\ref{fig:top}, we demonstrate that TDRL, one of the typical nonlinear ICA-based methods making the invertibility assumption, markedly degrades its performance in identifying the latent variables with increasing non-invertibility. It motivates us to extend the current nonlinear ICA methods to consider non-invertible mixing function. 
}

{In this paper, to tackle the challenges above, we propose to leverage the temporal context for retrieving missing information caused by the non-invertible mixing function, mirroring the intuitive mechanisms of human perception.
For instance, when we encounter an object with occlusion, our natural inclination is to draw from historical data to reconstruct the obscured portion.
We demonstrate that, even when the generation process is non-invertible, the derived latent causal representation remains identifiable if the latent variables can be expressed as an arbitrary function combining the current observation with its history.
Built upon this identification theorem, we introduce a principled approach, named \caring, that learns the function to integrate historical data to compensate for the latent information lost due to non-invertibility.
This approach extends the Sequential Variational Autoencoder (Sequential VAE~\citep{sequential_vae,li2018disentangled}) with two distinct modifications. Firstly, it incorporates history (or context) information directly into the encoder. 
Specifically, we transform step-to-step mapping (from current observation to the current latent variable) into sequence-to-step mapping (from current observation and temporal context to the current latent variable).
Secondly, a specialized prior module is introduced to determine the prior distribution of latent variables using the normalizing flow~\citep{dinh2016density}, ensuring the imposition of an independent noise condition. We evaluate our method using both synthetic and real-world data.}
Using synthetic data, we design datasets with a non-invertible mixing function to measure identifiability. For real-world applications, \ourmeos is deployed in a traffic accident reasoning task, a scenario in which the intricate traffic dynamics introduce considerable non-invertibility. Experimental outcomes reveal that our method significantly outperforms other temporal representation learning methods for identifying causal representations amid non-invertible generation processes. Furthermore, this causal representation has proven instrumental in enhancing video reasoning tasks.

\textbf{Key Insights and Contributions} of our research include:
\vspace{-0.1in}
\begin{itemize}
\item {To the best of our understanding, this paper presents the first identifiability theorem that accommodates a non-invertible generation process, which complements the existing body of the nonlinear ICA theory.}

\item {We present a principled approach, \caring, to learn the latent causal representation from temporal data under non-invertible generation processes with identifiability guarantees, by integrating temporal context information to recover the lost information.}

\item Our evaluations across synthetic and real-world datasets demonstrate the \caring's effectiveness for learning the identifiable latent causal representation, leading to enhancements in video reasoning tasks.
\end{itemize}
\vspace{-0.2in}

\section{Problem Setup}

\subsection{Non-invertible Temporal Generative Process}

{Denote $\mathbf{X} = \{\mathbf{x}_1,\mathbf{x}_2, \dots,\mathbf{x}_T\}$ as the observed $d$-dimensional} time series data at $T$ discrete time steps. 
Each observation $\mathbf{x}_t  \in \mathbb{R}^d$ is generated from a nonlinear mixing function $\mathbf{g}$ that maps $r+1$ adjacent latent variables $\mathbf{z}_{t:t-r} $ to $\mathbf{x}_t$, where $\mathbf{z}_{t:t-r}$ refers to 
$\{\mathbf{z}_{t},\mathbf{z}_{t-1},\cdots,\mathbf{z}_{t-r}\}$. We have $\mathbf{z}_t \in  \mathbb{R}^{n}$.  For every $i \in {1,\dots,n}$, the variable $z_{it}$ of $\mathbf{z}_t$ is derived from a stationary, non-parametric time-delayed causal relation:
\begin{equation}
\label{eq:generation}
\begin{aligned}
         &\underbrace{ \mathbf{x}_t = \mathbf{g}(\mathbf{z}_{t:t-r}) }_{\text{Nonlinear mixing}}, \\  &\underbrace{z_{it} = 
         {f_i\left(\{z_{j, t'} \vert z_{j, t'} \in \mathbf{Pa}(z_{it}) \}, \epsilon_{it}  \right)}}_{\text{Stationary non-parametric transition}}.
\end{aligned}
\end{equation}
Note that with non-parametric causal transitions, the noise term $\epsilon_{it} \sim p_{\epsilon_i}$ (where $p_{\epsilon_i}$ denotes the distribution of $\epsilon_{it}$) and the time-delayed parents $\mathbf{Pa}(z_{it})$ of $z_{it}$ (i.e., the set of latent factors that directly cause $z_{it}$)  are interacted and transformed in an arbitrarily nonlinear way to generate $z_{it}$. $\tau$ denotes the transition time lag.
The components of $\mathbf{z}_{t}$ are mutually independent conditional on history variables $\mathbf{Pa}(\mathbf{z}_{t})$.

{
In this case, one cannot recover $\mathbf{z}_t$ from $\mathbf{x}_t$ alone due to the non-invertibility of $\mathbf{g}$. Without extra assumptions, it is definitely non-identifiable. As a result, we assume that there exists a time lag $\mu$ and a nonlinear function $\mathbf{m}$ which can map a series of observations to latent variable $\mathbf{z}_t$, i.e.,
\begin{equation}
\label{eq:functionm}
\mathbf{z}_t = \mathbf{m}(\mathbf{x}_{t:t-\mu}).
\end{equation}
}

{
Once we successfully recover the information lost due to non-invertibility from the context, the classical nonlinear ICA algorithm can be used to solve this problem.
}

\subsection{Identification of the Latent Causal Processes}

\begin{definition}[Identifiable Latent Causal Process]
\label{def: 1}

Let $\mathbf{X} = \{\mathbf{x}_1, \mathbf{x}_2, \dots, \mathbf{x}_T\}$ be a sequence of observed variables generated by the true temporally causal latent processes specified by $(f_i, p({\epsilon_i}), \mathbf{g})$ given in Eq~\ref{eq:generation}. A learned generative model $(\hat{f}_i, \hat{p}({\epsilon_i}), \hat{\mathbf{g}})$ is observational equivalent to $(f_i, p({\epsilon_i}), \mathbf{g})$ if the model distribution $p_{\hat{f}_i, \hat{p}_\epsilon, \hat{\mathbf{g}}}(\mathbf{x}_{1:T})$ matches the data distribution $ p_{f_i, p_\epsilon,\mathbf{g}}(\mathbf{x}_{1:T})$ for any value of $\mathbf{x}_t$.
We say latent causal processes are identifiable if observational equivalence can lead to a version of latent variable $\mathbf{z}_t = \mathbf{m}(\mathbf{x}_{t:t-\mu})$ up to permutation $\pi$ and component-wise invertible transformation {$\mathcal{T}$}:
\begin{equation}\label{eq:iden}
\begin{aligned}
 &p_{\hat{f}_i, \hat{p}_{\epsilon_i},\hat{\mathbf{g}}}(\mathbf{x}_{1:T}) \! = \!  p_{f_i, p_{\epsilon_i}, \mathbf{g}} (\mathbf{x}_{1:T}) \\
 \Rightarrow & \hat{\mathbf{m}}(\mathbf{x}_{t:t-\mu}) \!= \!( {\mathcal{T}}  \! \circ \! \pi \!  \circ  \! \mathbf{m})(\mathbf{x}_{t:t-\mu}), 
 \forall \mathbf{x}_{t:t-\mu} \! \in \!\mathbb{R}^{\mu+1}.
\end{aligned}
\end{equation}
\end{definition} 

\vspace{-0.1in}

Different from the existing literature, {we involve $\mathbf{m}$ in the above definition since it serves implicitly as a property of the mixing function $\mathbf{g}$}, although it does not explicitly participate in the generation process.
Furthermore, the identifiability of $\mathbf{g}$ is different. In previous nonlinear ICA methods~\citep{yao2022temporally,hyvarinen2017nonlinear}, the mixing function $\mathbf{g}$ is identifiable.
However, in our case, we cannot find the identifiable mixing function since the information loss is caused by non-invertibility. Instead, we can obtain a component-wise transformation of a permuted version of latent variables $\mathbf{\hat{z}}_t = \mathbf{m}(\mathbf{x}_{t:t-\mu})$.
The latent causal relations are also identifiable, up to a permutation $\pi$ and component-wise invertible transformation $\mathcal{T}$, i.e., $\mathbf{\hat{f}} = \mathcal{T} \circ \pi \circ \mathbf{f}$, once $\mathbf{z}_t$ is identifiable.
Because in the time-delayed causally sufficient system, the conditional independence relations fully characterize time-delayed causal relations when we assume no latent causal confounders in the (latent) causal processes.
\vspace{-0.1in}

\begin{figure}[t]
    \centering
    \includegraphics[width=0.95\linewidth]{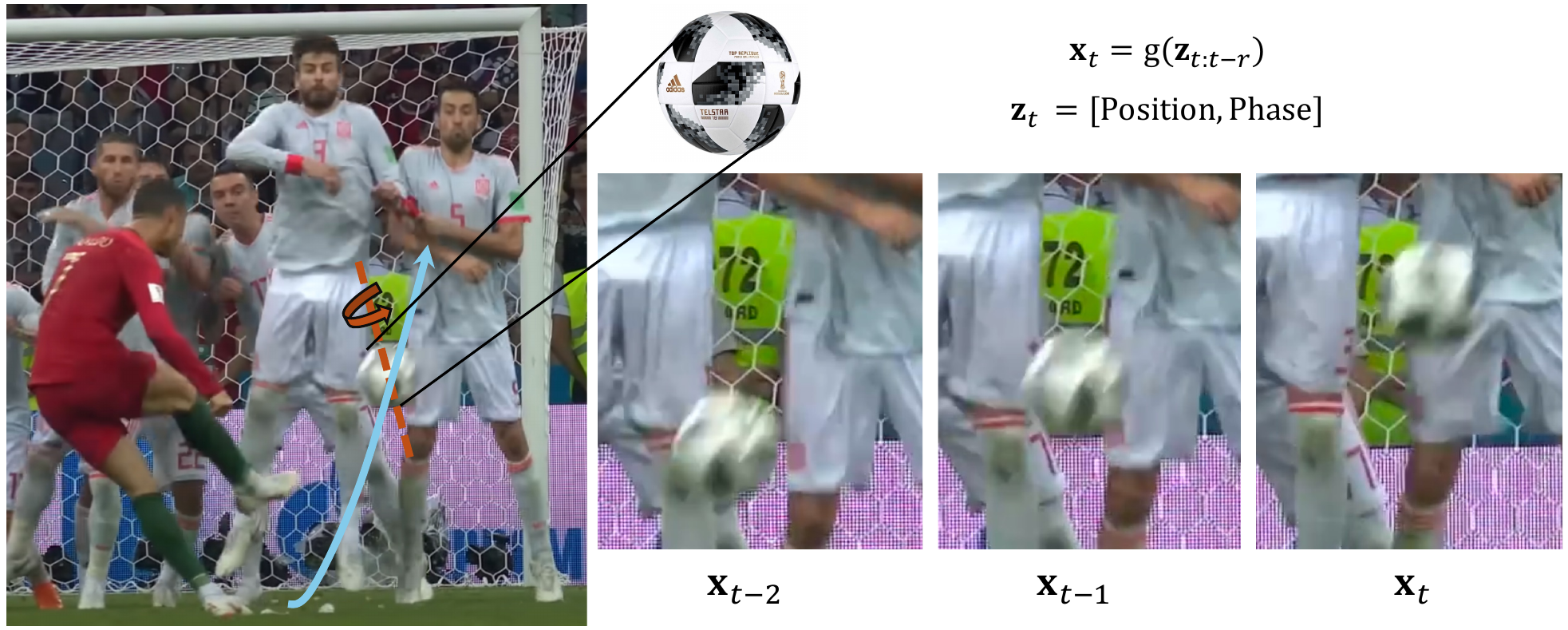}
 
    \caption{
        {
            {
            \textbf{An intuitive illustration of a moving football with a visual persistence effect.} Considering the generating process $\mathbf{x}_t=\mathbf{g}(\mathbf{z}_{t:t-r})$, $\mathbf{x}_t$ denotes the observed football with motion blur, and $\mathbf{z}_t$ denotes the position and phase of the ball. Recovering the latent variables from a single observation will be difficult, which introduces non-invertibility.
            }
        }
    }
    \vspace{-0.4cm}
    \label{fig: toy-example}
\end{figure}

\subsection{Illustrations of the Problem Setup}

\textbf{Intuitive Illustration with Visual Persistence.}
Consider a rapidly moving ball on a two-dimensional plane as described in figure~\ref{fig: toy-example}. The horizontal and vertical coordinates of the ball's position at any given moment can be represented by the latent variable $\mathbf{z}_t\in \mathbb{R}^2$. We assume that the ball follows a curved trajectory constrained by the nonlinear function $\mathbf{f}$  as it moves.

Suppose that we observe the ball with a visual persistence effect, where each observation $\mathbf{x}_t$ captures several consecutive latent variables as 
$\mathbf{x}_t = \mathbf{g}(\mathbf{z}_{<t})$.
The mixing function $\mathbf{g}$ refers to the weighted sum of the images obtained through multiple exposures, which is what a person ultimately observes as $\mathbf{x}_t$. In this case, the invertibility of the mapping from $\mathbf{z}_{t}$ to $\mathbf{x}_{t}$ is compromised since the current frame also contains the latent information from previous frames.

\textbf{Mathematical Illustration.}
Besides, we provide a mathematical example to demonstrate the existence of function $\mathbf{m}$ in Eq~\ref{eq:functionm}. Following the concept of visual persistence, let the current observation be a weakened previous observation overlaid with the current image of the object, i.e., $\mathbf{x}_t = \mathbf{z}_t + \frac{1}{2}\mathbf{x}_{t-1} = \sum_{i=1}^{\infty} \left(\frac{1}{2}\right)^{i} \mathbf{z}_{t-i}$ \citep{wolford1993model}. 
Given an extra observation, the current latent variable can be rewritten as $\mathbf{z}_t  = \mathbf{x}_t - \frac{1}{2}\mathbf{x}_{t-1}$.
Thereby we can easily recover latent variables that cannot be obtained from a single observation, i.e., $\mathbf{z}_t = \mathbf{m}(\mathbf{x}_{t:t-1}) = \mathbf{x}_t - \frac{1}{2}\mathbf{x}_{t-1}$.

\textbf{Illustration of time-delayed temporal relations.}
Here, we assume that there are only time-delayed temporal relations in the time series system. In other words, any instantaneous relations will not fall into the discussion. Generally speaking, a group of objects that have instantaneous relations with each other would be treated as one single variable. 
For example, within a video sequence, a ball in motion may be conceptualized as a cluster of pixels that move consistently and simultaneously (instantaneous relations). This pattern can help distinguish the ball from the others, which potentially provides a principle to extract concepts from time series data like video, motion sequence, etc.

\section{Identifiability Theory}\label{sec:theory}

{
In this section, we demonstrate that, given certain mild conditions, the learned causal representation $\mathbf{z}_t$ is identifiable up to permutation and a component-wise transformation. This holds even if the mixing function $\mathbf{g}$ is non-invertible. Firstly, we present the identifiability results when faced with a non-invertible mixing function and stationary transitions. Subsequently, we address the gap between permutation-scaling Jacobian and identifiability. Lastly, by leveraging side information such as the domain index and label, we illustrate how identifiability can be achieved even in a non-stationary context. The 
proofs are available in Appendix~\ref{ap:theory}.
}

\subsection{Identifiability under Non-Invertible Generative Process}

W.L.O.G., we first consider a simplified case with $\tau=r+1$ and context length $\mu$, which infers such process:
\begin{equation}
    \label{eq:process_lag1}
   \mathbf{x}_t = \mathbf{g}(\mathbf{z}_{t:t-r}) , \quad z_{it} = f_i\left(\mathbf{z}_{t-1:t-r-1}, \epsilon_{it}  \right),
\end{equation}
where a function $\mathbf{m}$ satisfying $\mathbf{z}_t = \mathbf{m}(\mathbf{x}_{t:t-\mu})$ exists. When taking $r=0$, the time delay is present only in transitions and is absent in the generation process. Taking $r>0$ leads us to a more intricate scenario, where the mixing function encompasses not just the latent causal variables of the current time step, but also the information of previous steps, termed the Time-delayed Mixing Process. Such a scenario is compelling, acknowledging that the mixing process can be influenced by time-delayed effects. To illustrate, human visual perception provides a fitting example: the phenomenon known as the persistence of vision reveals that humans retain impressions of a visual stimulus even after its cessation~\citep{coltheart1980persistences}. The extensions for any time lag $\tau$ will be discussed in Appendix~\ref{ap:moretau}.

\begin{theorem}
[Identifiability under Non-invertible Generative Process] \label{Th1} 

For a series of observations {$\mathbf{x}_t\in\mathbb{R}^d$} and estimated latent variables $\mathbf{\hat{z}}_{t}\in\mathbb{R}^n$, suppose there exists function $\mathbf{\hat{g}},\mathbf{\hat{m}}$ which is subject to observational equivalence, 
\begin{equation}
    \label{eq:invert_lag1}
    \mathbf{x}_t = \mathbf{\hat{g}}(\mathbf{\hat{z}}_{t:t-r}),\quad \mathbf{\hat{z}}_{t} =\mathbf{\hat{m}}(\mathbf{x}_{t:t-\mu}).
\end{equation}
If assumptions
\begin{itemize}
    \item {(Smooth and Positive Density) the probability density function of latent variables is third-order differentiable and positive in $R^n$,}
    \item (conditional independence) the components of $\mathbf{\hat{z}}_{t}$ are mutually independent conditional on $\mathbf{\hat{z}}_{t-1:t-r-1}$,
    {\item (sufficiency) let $\eta_{kt} \triangleq \log p(z_{kt} | \mathbf{z}_{t-1:t-r-1})$, and
\begin{equation} \label{Eq:v}
\begin{aligned}
\mathbf{v}_{lt} 
\triangleq \Big(
\frac{\partial^2 \eta_{1t}}{\partial z_{1t} \partial z_{l,t-r-1}}, 
..., 
\frac{\partial^2 \eta_{nt}}{\partial z_{nt} \partial z_{l,t-r-1}}, \\
\frac{\partial^3 \eta_{1t}}{\partial z_{1t}^2 \partial z_{l,t-r-1}}, 
..., 
\frac{\partial^3 \eta_{nt}}{\partial z_{nt}^2 \partial z_{l,t-r-1}} \Big)^\intercal,
\end{aligned}
\end{equation}
for $l=1,2,\cdots,n$. For each value of $\mathbf{z}_t$, there exists $2n$ different values of $z_{l,t-r-1}$ such that the $2n$ vector functions $\mathbf{v}_{lt}\in\mathbf{R}^{2n}$ are linearly independent,}
\end{itemize}
are satisfied, then $\mathbf{z}_{t}$ must be a component-wise transformation of a permuted version of $\mathbf{\hat{z}}_{t}$  with regard to context $\{\mathbf{x}_j\mid\forall j=t,t-1,\cdots,t-\mu-r\}$.
\end{theorem}

{
The proof of Theorem~\ref{Th1} can be found in Appendix~\ref{sec: app_th1}. It is inspired by \cite{yao2022temporally}, which follows the line of \cite{hyvarinen2019nonlinear}.
}

{Besides, the nonstationary transition can also help to improve the identifiability of \caring. As shown in the sufficiency assumption in Theorem~\ref{Th1}, the identifiability relies on the sufficient changes of the conditional distribution $p(z_{kt}|\mathbf{z}_{t-1:t-r-1})$. When the distribution of the noise term varies between different domains, the domain index can serve as an auxiliary variable to improve this sufficiency since both domain dynamics and history variables can provide changes. More discussions are in Appendix \ref{sec: non-stationary}.}

\subsection{Continuity for Permutation Invariance}

{
In this subsection, we will introduce permutation invariance for further discussion.
}

\begin{definition}[Permutation Invariance]
{
{Following Definition ~\ref{def: 1}, if $\pi$ is a fixed permutation and $\mathcal{T}$ is a component-wise invertible transformation which may vary across different time steps, we call this identifiability under Permutation Invariance.}
}    
\end{definition}

Let us further consider a more general scenario, with $\mathbf{x}_t\in\mathcal{X}\subseteq\mathbb{R}^d$ and $\mathbf{z}_t\in\mathcal{Z}\subseteq\mathbb{R}^n$,
i.e., the {probability density} of $\mathbf{z}_t$ does not have to be non-zero everywhere in $\mathbb{R}^n$ .
To establish identifiability, numerous existing nonlinear ICA-based methods~\citep{yao2021learning,yao2022temporally,hyvarinen2019nonlinear,halva2021disentangling} utilize the Jacobian matrix, denoted by $\mathbf{H}=\frac{\partial \mathbf{z}}{\partial \hat{\mathbf{z} } }$, which captures the relationship between ground truth and estimated latent variables. These methods propose that the learned latent variables are identifiable if $\mathbf{H}_{ij}\cdot\mathbf{H}_{ik}=0$ for $j\not=k$ (with only a single non-zero element in each row or column). $\mathbf{H}$ corresponds to the Jacobian matrix of the function $\mathbf{h} \triangleq \mathbf{m}\circ\hat{\mathbf{g}}$ in our scenario (or $\mathbf{g}^{-1}\circ\hat{\mathbf{g}}$ for the general scenario). However, it is crucial to highlight an often overlooked shortcoming: this condition alone is insufficient to establish identifiability when dealing with non-linear generation processes. Concurrently to our work, \cite{lachapelle2023additive} also arrived at the difference between local and global disentanglement, and achieved the global disentanglement under the additive decoding case. Alternatively, we demonstrate the identifiability under the permutation invariance and focus on a more general case without the block-specific decoder assumptions.
While in linear ICA, given that the Jacobian remains constant, this condition indeed equates to identifiability. Yet, in nonlinear ICA, the Jacobian matrix, being a function of $\mathbf{\hat{z}}$, can vary with different $\mathbf{\hat{z}}$ values, potentially rendering the mapping unpredictable. 
A comprehensive discussion is available in Appendix~\ref{ap: necessity of continuity}.

To solve this issue in the nonlinear system, we provide two more assumptions. The domain $\mathcal{\hat{Z}}$ of $\mathbf{\hat{z}}$ should be path-connected, i.e., for any $\mathbf{a}, \mathbf{b} \in \mathcal{\hat{Z}}$, there exists a continuous path connecting $\mathbf{a}$ and  $\mathbf{b}$ with all points of the path in $\mathcal{\hat{Z}}$. In addition, function $\mathbf{h}$ is second-order differentiable and holds the non-degeneracy condition.

For clarification, the condition that a function $\mathbf{h}:\mathbb{R}^n\rightarrow\mathbb{R}^n$ is invertible, or equivalently the non-vanishing of the determinant of the Jacobian matrix $\mathbf{H}_h$, is called the non-degeneracy condition. We first define the partially invertible function, and then give the non-degeneracy condition on it.

\begin{definition}[Partially Invertiblility]
    {
        A function $\mathbf{z}=\mathbf{h}(\mathbf{\hat{z}},\mathbf{c})$, where $\mathbf{z},\mathbf{\hat{z}}\in\mathbb{R}^n$ and $\mathbf{c}\in\mathbb{R}^m$, is partially invertible, if and only if for any given $\mathbf{c}$, the rest part $\mathbf{h}_\mathbf{c}:\mathbb{R}^n\rightarrow\mathbb{R}^n$ is always invertible.
    }
\end{definition}

\begin{definition}[Non-degeneracy Condition of Partially Invertible Functions]
{
    The non-degeneracy condition of a partially invertible function $\mathbf{z}=\mathbf{h}(\mathbf{\hat{z}},\mathbf{c})$ is that for any given $\mathbf{c}$, the determinant of the Jacobian matrix $\mathbf{H}_{\mathbf{h}_\mathbf{c}}$ of $\mathbf{h}_\mathbf{c}$ is always non-zero.
    }
\end{definition}

{
\begin{lemma}[Disentanglement with Continuity]
    \label{lemma: continuousity}
    For second-order differentiable invertible function $\mathbf{h}$ defined on a path-connected domain $\mathcal{\hat{Z}}\subseteq \mathbb{R}^n$ which satisfies $\mathbf{z} = \mathbf{h}(\mathbf{\hat{z}})$, suppose the non-degeneracy condition holds. If there exists at most one non-zero entry in each row of the Jacobian matrix $\mathbf{H}=\frac{\partial \mathbf{z}}{\partial \mathbf{\hat{z}}}$, the identifiability under Permutation Invariance can be established.
\end{lemma}
}

Furthermore, when the Jacobian matrix is more than a function of $\mathbf{\hat{z}}$, but also is influenced by a side information $\mathbf{c}$, the identifiability can be guaranteed under mild extra conditions. 

{
\begin{lemma}[Disentanglement with Continuity under Side Information]
    \label{lemma: coutinuous ++}
    For second-order differentiable invertible function $\mathbf{h}$ defined on a path-connected domain $\mathcal{\hat{Z}} \times \mathcal{C}\subseteq \mathbb{R}^{n+m}$ which satisfies $\mathbf{z} = \mathbf{h}(\mathbf{\hat{z}},\mathbf{c})$, suppose the non-degeneracy condition holds. If there exists at most one non-zero entry in each row of the Jacobian matrix $\mathbf{H}(\mathbf{c})=\frac{\partial \mathbf{z}}{\partial \mathbf{\hat{z}}}$, the identifiability under Permutation Invariance can be established.
\end{lemma}
}

With Lemma~\ref{lemma: coutinuous ++}, we can further extend Theorem~\ref{Th1} to guarantee permutation invariance even when the probability density of $\mathbf{z}$ is not positive everywhere on $\mathbb{R}^n$, as long as appropriate continuity conditions are satisfied. This serves as a valuable complement to the existing theory of nonlinear ICA, which further relaxes the required assumptions. This relaxation enhances the robustness of \ourmeos and makes it more adaptable to diverse and complex data, thus improving its applicability in practical settings.

\begin{proposition}\label{proposition} 

For a series of observations $\mathbf{x}_t\in\mathcal{X}\subseteq\mathbb{R}^d$ and estimated latent variables $\mathbf{\hat{z}}_{t}\in\mathcal{Z}\subseteq\mathbb{R}^n$, suppose there exists function $\mathbf{\hat{g}},\mathbf{\hat{m}}$ which subject to observational equivalence, i.e.,
\begin{equation}
    \label{eq:invert_lagn}
    \mathbf{x}_t = \mathbf{\hat{g}}(\mathbf{\hat{z}}_{t:t-r}),\quad \mathbf{\hat{z}}_{t} =\mathbf{\hat{m}}(\mathbf{x}_{t:t-\mu}).
\end{equation}
where $\mathbf{g},\mathbf{\hat{g}},\mathbf{m},\mathbf{\hat{m}}$ are second-order differentiable.
In addition, if assumptions the same as Theorem~\ref{Th1} are satisfied, then the identifiability of $\mathbf{z}_{t}$ under Permutation Invariance can be established. 
\end{proposition}

\begin{figure}[t]
    \centering
    \vspace{0cm}
 
        \includegraphics[width=\linewidth]{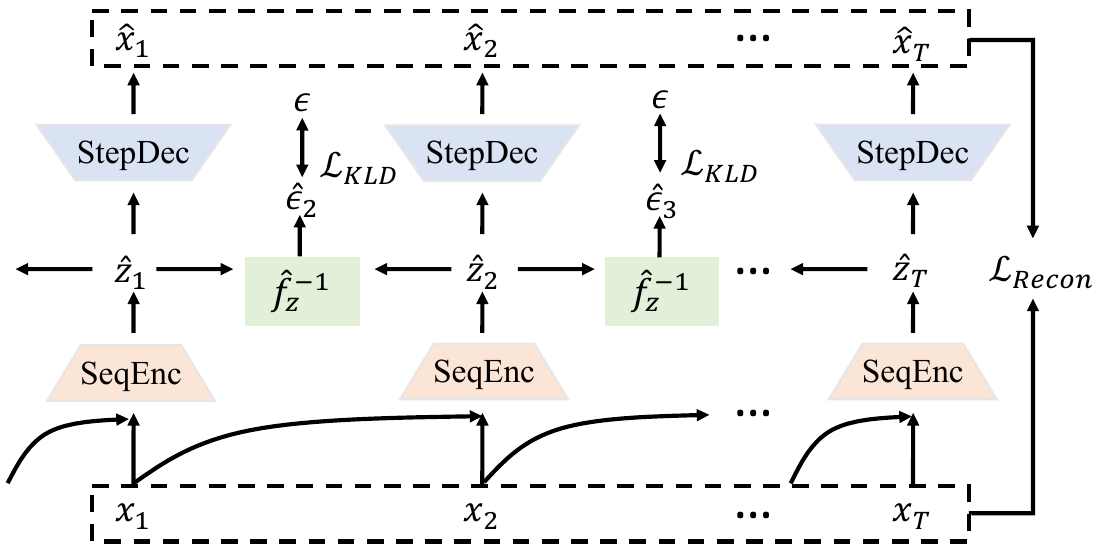}

        \caption{\textbf{The overall framework of \caring.} It consists of three main modules, including the sequence-to-step encoder, step-to-step decoder, and the transition prior module, which is represented as $\text{SeqEnc}$, $\text{StepDec}$, and $\hat{\mathbf{f}}^{-1}_{\mathbf{z}}$ in a different color, respectively. The model is trained with both $\mathcal{L}_{Recon}$ and $\mathcal{L}_{KLD}$.}
        \label{fig:framework}

    \vspace{-0.2in}
\end{figure}

\section{Approach}
\label{sec:method}
{
Given our results on identifiability, we introduce our \ourmeos approach. This aims to estimate the latent causal dynamics presented in Eq~\ref{eq:generation}, even when faced with a non-invertible mixing procedure. 
To achieve this, \ourmeos builds upon the Sequential Variational Auto-Encoders (Sequential VAE  \citep{sequential_vae,li2018disentangled}) and incorporates three primary modules: the sequence-to-step encoder (SeqEnc), the step-to-step decoder (StepDec), and the transition prior module ($\hat{\mathbf{f}}^{-1}_{\mathbf{z}}$). Through Sequential VAE, we ensure the reconstruction capability from latent variables to observed variables. Meanwhile, in contrast to the Gaussian prior in VAEs, our method employs normalizing flow to control the prior distribution, ensuring that the latent variables satisfy the assumed conditional independence. During the training phase, we integrate the conditions from Sec.~\ref{sec:theory} as constraints and adopt two corresponding loss functions.
}

\bfsection{Overall Framework} As visualized in Figure~\ref{fig:framework}, our framework starts by acquiring the latent causal representation via a sequence-to-step encoder, whose input and output are a sequence of observations $\mathbf{x}_{t:t-\mu}$ and the estimated latent variable $\mathbf{\hat{z}}_t$. Formally, it denotes the inference process of $q(\mathbf{\hat{z}}_t|\mathbf{x}_{t:t-\mu})$, which corresponds to the function $\mathbf{m}$ in Eq~\ref{eq:functionm}.  
Following this, observations are generated from the latent space through a step-to-step decoder $p(\mathbf{\hat{x}}_t|\mathbf{\hat{z}}_t)$, which implies the mixing function $\mathbf{g}$ as mentioned in Eq~\ref{eq:generation}. 
To learn the independent latent variables, we apply a constraint using the KL divergence between the posterior distribution of learned latent variables and a prior distribution which is subject to our conditional independence assumption in Theorem~\ref{Th1}. 
The estimation of the prior distribution motivates us to utilize a normalizing flow, converting the prior distribution into Gaussian noise, represented as $\hat{\epsilon}_{it} = f^{-1}_i(\hat{z}_{it},\hat{\mathbf{z}}_{t-1:t-\tau})$. 
Moreover, a reconstruction loss between the ground truth and generated observations is integrated for model training.
A detailed exploration of all modules and losses is forthcoming.

\bfsection{Sequence-to-Step Encoder and Step-to-Step Decoder}
Drawing inspiration from the capability of the human visual system, we utilize temporal context to reclaim the information lost due to non-invertible generation. The human visual system adeptly fills in occluded segments by recognizing coherent motion cues~\citep{palmer1999vision,wertheimer1938laws,spelke1990principles}. 
Assuming there's a function that captures all latent information from the current observation and its temporal context, we can retrieve the latent causal process with identifiability, i.e. $\mathbf{m}$ exists.
Various non-linear models are suitable for estimating this function, taking a sequence of observations, $\mathbf{x}_{t:t-\mu}$, with a lag of $\mu$ as inputs, and yielding the estimated latent representation of the current time step as output. In our experiments, we utilize both Multi-Layer Perceptron (MLP)~\citep{werbos1974beyond} and Transformer~\citep{vaswani2017attention}, catering to different complexities.
Given the estimated latent variable $\hat{\mathbf{z}}_t$, a step-to-step decoder is employed to generate the current observation $\mathbf{x}_t$. For practical implementation, one MLP is sufficient.

\bfsection{Transition Prior Module} 
{
To uphold the conditional independence assumption, we propose to minimize the KL divergence between the posterior distribution and a hard-coding prior distribution with such property. 
The constraint indicates that current latent variables are mutually independent, conditioned on historical latent variables. Formally, by hard-coding the prior distribution we enforce $\mathbf{\hat{z}}_ {t} | \mathbf{\hat{z}}_ {t-1:t-\tau}$ to be mutually independent. By minimizing the KL divergence, we expect the posterior to be subject to the assumption as well, such as $\mathbf{\hat{z}}_ t |\mathbf{\hat{x}}_ {t:t-\mu},\mathbf{\hat{z}}_ {t-1:t-\tau}$ are mutually independent.
}
Direct estimation of the prior, which has an arbitrary density function, poses challenges. As a solution, we introduce a transition prior module that facilitates the estimation of the prior using normalizing flow. Specifically, the prior is represented through a Gaussian distribution combined with the Jacobian matrix of the transition module.

Formally presented, the transition prior module is represented as $\hat{\epsilon}_{it}=\hat{f}^{-1}_i(\hat{z}_{it},\mathbf{\hat{z}}_{t-1:t-\tau})$.  
Subsequently, the joint distribution is decomposed as a product of the noise distribution and the determinant of the Jacobian matrix, formulated as $p([\mathbf{\hat{z}}_{t-1:t-\tau}, \mathbf{\hat{z}}_t]) = p([\mathbf{\hat{z}}_{t-1:t-\tau}, \mathbf{\hat{\epsilon}}_t]) \times |\mathbf{J}| $, with 
{
$\mathbf{J} = \begin{bmatrix}
        \mathbb{I}_{n\tau} & \mathbf{0} \\
        \mathbf{0} & diag(\frac{\partial \hat{\epsilon}_{it}}{\partial \hat{z}_{it}})
    \end{bmatrix}$,
}
where $[\cdot]$ denotes concatenation. Leveraging this joint distribution, we derive the prior as
\begin{equation}
\label{eq:prior}
\begin{aligned}
   \quad & \log p(\mathbf{\hat{z}}_t|\mathbf{\hat{z}}_{t-1:t-\tau}) \\
    & = \log p([\mathbf{\hat{z}}_t,\mathbf{\hat{z}}_{t-1:t-\tau}]) - \log p(\mathbf{\hat{z}}_{t-1:t-\tau}) \\
    & = \log p([\mathbf{\hat{\epsilon}}_t,\mathbf{\hat{z}}_{t-1:t-\tau}]) + \log|\mathbf{J}| - \log p(\mathbf{\hat{z}}_{t-1:t-\tau}) \\
    & = \log p(\mathbf{\hat{\epsilon}}_t|\mathbf{\hat{z}}_{t-1:t-\tau}) + \log|\mathbf{J}| \\
    & = \log p(\mathbf{\hat{\epsilon}}_t) + \log|\mathbf{J}| \\
    & = \sum_i\log p(\hat{\epsilon}_{it}) + \log|\mathbf{J}| \quad \text{: Conditional independence}\\
    & = \sum_i\left(\log p(\hat{\epsilon}_{it}) + \log \frac{\partial \hat{\epsilon}_{it}}{\partial \hat{z}_{t,i}}\right)\text{: Lower-triangular}.
\end{aligned}
\end{equation}
The transition prior module can be efficiently executed using an MLP, transforming the latent variables $\mathbf{\hat{z}}_{t:t-\tau}$ into $\mathbf{\hat{\epsilon}}_{t}$.

 \begin{figure*}[t]
\begin{center}
\centerline{\includegraphics[width=1.0\linewidth]{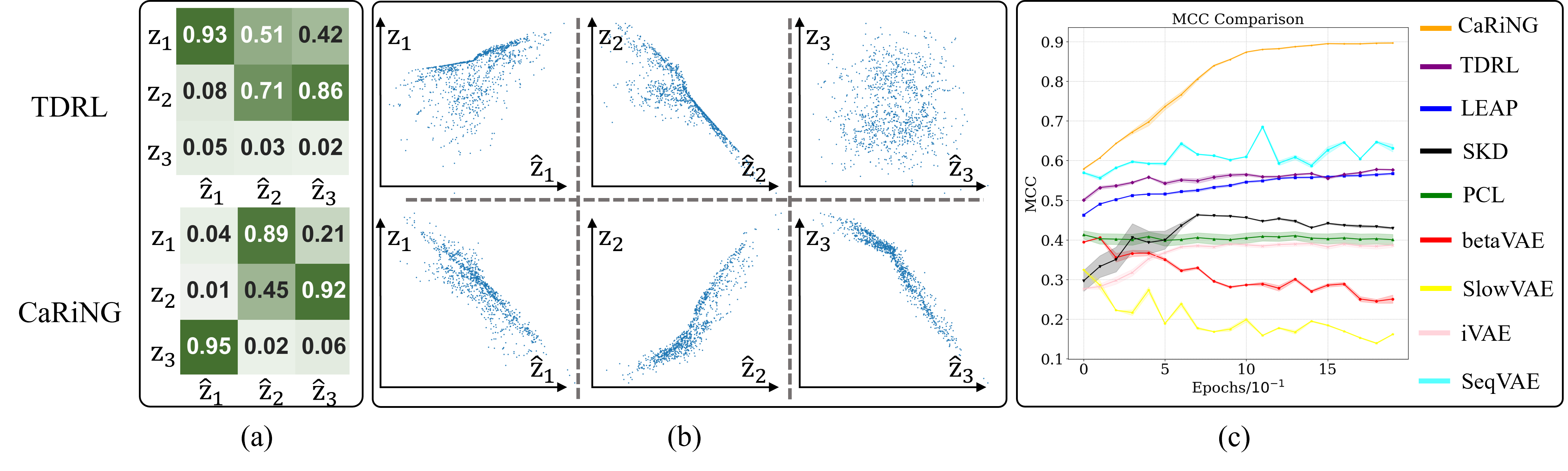}}
\vspace{-0.1in}
\caption{{\textbf{Qualitative comparisons between baselines (especially TDRL) and \ourmeos in the setting of Non-invertible Generation.} (a) MCC matrix for all 3 latent variables; (b) The scatter plots between the estimated and ground-truth latent variables (only the aligned variables are plot); (c) The validation MCC curves of \ourmeos and other baselines. }}
\label{fig:exp}
\end{center}
\vspace{-0.2in}
\end{figure*}

\bfsection{Optimization} We train \ourmeos using the Evidence Lower BOund (\textsc{ELBO}) objective, which is written as follows:
\begin{equation}
\small
\begin{split}
&\textsc{ELBO}\!\triangleq \! \mathbb{E}_{q_{\phi}(\mathbf{Z}|\mathbf{X})}[\log p_{\theta}(\mathbf{X}|\mathbf{Z}) ] - D_{KL}(q_{\phi}(\mathbf{Z}\vert \mathbf{X})\vert\vert p_{\theta}(\mathbf{Z}))\\
&\!=\!\underbrace{\mathbb{E}_{q_{\phi}(\mathbf{Z}|\mathbf{X})}\sum_{t=1}^{T} \log p_{\theta}(\mathbf{x}_t\vert\mathbf{z}_t)}_{-\mathcal{L}_{\text{Recon}}}  \\
&\!+\!\underbrace{\mathbb{E}_{q_{\phi}(\mathbf{Z}|\mathbf{X})}\left[ \sum_{t=1}^{T}\log  p_{\theta}(\mathbf{z}_t\vert \mathbf{z}_{t-1:t-\tau}) 
\!-\!\sum_{t=1}^{T}\log q_{\phi}(\mathbf{z}_t\vert\mathbf{x}_{t:t-\mu})\right]}_{-\mathcal{L}_{\text{KLD}}}. 
\end{split}
\end{equation}
For the reconstruction likelihood $\mathcal{L}_{\text{Recon}}$, we utilize the mean-squared error (MSE) to measure the discrepancy between the generated and original observations. When computing the KL divergence $\mathcal{L}_{\text{KLD}}$, we resort to a sampling method, given that the prior distribution lacks an explicit form. To elaborate, the posterior is produced by the encoder, while the prior is defined as in Eq~\ref{eq:prior}.

\begin{table*}[t]
\centering

\caption{\textbf{MCC scores (with standard deviations over 4 seeds) of \ourmeos and baselines on NG and NG-TDMP settings.} }
\vspace{-0.1cm}
\label{tab:syn-results}
\resizebox{\textwidth}{!}{%
\begin{tabular}{@{}l|ccccccccc@{}}
\toprule
\multirow{2}{*}{\textbf{\begin{tabular}[c]{@{}c@{}}Setting\end{tabular}}} & \multicolumn{9}{c}{\textbf{Method}}                               \\ \cmidrule(l){2-10} 
& \ourmeos & TDRL & LEAP & SlowVAE & PCL  & betaVAE & SKD & iVAE & SequentialVAE\\ \midrule
NG &
  \textbf{0.933 \textpm 0.010} &
  0.627 \textpm 0.009 &
  0.651 \textpm 0.019 &
  0.362 \textpm 0.041 &
  0.507 \textpm 0.091 &
  0.551 \textpm 0.007 &
  0.489 \textpm 0.077 &
  0.391 \textpm 0.686 &
  0.750 \textpm 0.035
   \\
NG-TDMP &
  \textbf{0.921 \textpm 0.010} &
  0.837 \textpm 0.068 &
  0.704 \textpm 0.005 &
  0.398 \textpm 0.037 &
  0.489 \textpm 0.095 &
  0.437 \textpm 0.021 &
  0.381 \textpm 0.084 &
  0.553 \textpm 0.097 &
  0.847 \textpm 0.019
   \\
   \bottomrule
\end{tabular}%
}
\vspace{-0.1in}
\end{table*}

\section{Experiments}

\label{sec:exp}
We conducted the experiments in two simulated environments, utilizing the available ground truth latent variables to evaluate identifiability. Subsequently, we assessed \ourmeos on a real-world VideoQA task, SUTD-TrafficQA~\citep{trafficqa}, to verify its capability in representing complex and non-invertible traffic events.

\subsection{Simulation Experiments}

{
\bfsection{Dataset and experimental settings} To evaluate whether \ourmeos can learn the causal process and identify the latent variables under a non-invertible scenario, we design a series of simulation experiments based on a random causal structure with a given sample size and variable size. We provide two experimental settings, including $\text{NG}$ and $\text{NG-TDMP}$, which simulate the scenarios in Theorem~\ref{Th1} with $r=0$ (non-invertible generation) and $r>0$ (time-delayed mixing process), respectively. 
In particular, for $\text{NG}$, we simulate the visual perception system that uses the ground-truth dimension as 3 to represent the 3D real world and apply 2 measured variables to represent the 2D observation, which indicates the generation is non-invertible. 
For $\text{NG-TDMP}$, we simulate the persistence of vision that involves the previous latent variables in the current mixing process. It denotes that even if the dimension of the observation is not reduced, the generation process is still non-invertible due to the time-delay mixing.  
More details of the data generation process can be found in Appendix~\ref{ap:synthetic}. 
}

\bfsection{Evaluation metrics} We apply the standard evaluation metric in the field of ICA, Mean Correlation Coefficient (MCC), to evaluate the identifiability of our \caring. MCC measures the recovery of latent factors by calculating the absolute values of the correlation coefficient between every ground-truth factor against
every estimated latent variable. It first calculates the Pearson correlation coefficients to measure the relationship and then adjusts the order with an assignment algorithm.
The MCC score is a value from 0 to 1, where the higher score denotes better identifiability.

{
\bfsection{Baseline methods} We compare \ourmeos with a series of baseline methods. BetaVAE~\citep{higgins2016beta} is the most basic baseline which ignores the temporal dependency and cannot utilize any auxiliary information. 
SlowVAE~\citep{klindt2020towards}, and PCL~\citep{hyvarinen2017nonlinear} show the identifiability results but are limited by the assumption of independent sources. 
iVAE~\citep{khemakhem2020variational} leverage nonstationarity (auxiliary information) to achieve identifiability.
It is important to note that iVAE requires additional domain labels as input. In our experiments, we simply used time indices as the domain label. 
In addition, LEAP~\citep{yao2021learning} and TDRL~\citep{yao2022temporally} allow for learning causal processes but assume an invertible generation process. 
Besides, we also compare \ourmeos with other temporal representation learning methods that are not based on ICA, such as Sequential VAE~\citep{sequential_vae} and SKD~\citep{berman2022multifactor}, in which the disentangled representation has no identifiability guarantee.
}

\begin{table}[t]\renewcommand\tabcolsep{7.0pt}\renewcommand\arraystretch{0.95}
\begin{center}
\caption{\textbf{Results on the SUTD-TrafficQA dataset.} The cross-modality matching parts of TDRL and \ourmeos are based on HCRN. }
\vspace{-0.1cm}
\label{tab: trafficqa} 
\begin{tabular}{l|c|c}
\toprule
Method & Year & Accuracy($\%$) \\
\midrule
I3D+LSTM&CVPR2017&33.21\\
HCRN &CVPR2020&36.26\\
VQAC &ICCV2021&36.00\\
MASN &ACL2021&36.03\\
DualVGR&TMM2021&36.07\\
Eclipse &CVPR2021&37.05\\
CMCIR &TPAMI2023 &38.58\\
\midrule
TDRL&NeurIPS2022&37.32\\
\ourmeos& - &\textbf{41.22}\\
\bottomrule
\end{tabular}
\end{center}
\vspace{-0.5cm}
\end{table}

\bfsection{Quantitative results} 
The performance of \ourmeos and other baseline methods in both the NG and NG-TDMP scenarios is presented in Table~\ref{tab:syn-results}. Initially, it's evident that all baseline Nonlinear ICA methods yield unsatisfactory MCC scores in both scenarios, including the strong TDRL baseline, which previously obtained good results in invertible settings, as shown in Figure~\ref{fig:exp} (c). 
As shown in Figure~\ref{fig:exp} (a), TDRL cannot recover the lost latent variables caused by non-invertible generation (MCC=0.03 for that variable). 
It is also illustrated by the scatter plots in Figure~\ref{fig:exp} (b), which show the independence between the estimated and ground truth variables on that dimension. 
Interestingly, we find that the Sequential VAE method works better than other methods that don't use the temporal context, which also demonstrates the necessity of temporal context to solve the invertibility issue. 
However, we still find that constraining the conditional independence benefits better performance, which shows the effect of the KL part.
Furthermore, \ourmeos consistently delivers robust identifiability outcomes in both settings. This suggests that leveraging temporal context significantly enhances identifiability when faced with non-invertible generation processes. Lastly, performance in the NG scenario is better than that in the NG-TDMP scenario, showing the increased complexity introduced by the time-delayed mixing process.

\subsection{Real-world Experiments}

\bfsection{Dataset and experimental settings}
The SUTD-TrafficQA dataset~\citep{trafficqa} is a comprehensive resource tailored for video event understanding in traffic scenarios, notably characterized by numerous occlusions among traffic agents. 
 It consists of 10,090 videos and provides over 62,535 human-annotated QA pairs.
Among them, 56,460 QA pairs are used for training and the rest 6,075 QA pairs are used for testing.
The dataset challenges models with six reasoning tasks: ``Basic Understanding'' is designed for grasping traffic dynamics. ``Event Forecasting'' and ``Reverse Reasoning'' evaluate the temporal prediction ability. 
``Introspection'', ``Attribution'', and ``Counterfactual Inference'' require the model to understand the causal dynamic and conduct reasoning. 
All tasks are formulated as multiple-choice forms (evaluation with accuracy) without limiting the number of candidate answers, and demand a deep comprehension of traffic events and their underlying causality. 

\bfsection{Baseline methods} 
The primary method we benchmark against is TDRL~\citep{yao2022temporally},
to evaluate the representation ability of the complex and non-invertible traffic environment.  Additionally, we evaluate \ourmeos in comparison with state-of-the-art VideoQA methods, including I3D+LSTM~\citep{carreira2017quo}, HCRN~\citep{le2020hierarchical}, VQAC~\citep{kim2021video}, MASN~\citep{seo2021attend}, DualVGR~\citep{wang2021dualvgr}, Eclipse~\citep{trafficqa}, and CMCIR~\citep{liu2023cross}. In our approach, \ourmeos is leveraged to identify latent causal dynamics, while HCRN serves as the basic model for question answering. Further implementation details are provided in the Appendix.

\bfsection{Quantitative results} 
Performance comparisons for the six question types on SUTD-TrafficQA are summarized in Table~\ref{tab: trafficqa}.
\ourmeos achieves a score of 41.22, which demonstrates a significant improvement which is nearly 6.8\% over the next best method.
Notably, when compared to TDRL, which lacks temporal context, \ourmeos exhibits significant advancements in representing complex, non-invertible traffic events.
When benchmarked against the HCRN baseline, which employs the same cross-modality matching module, our approach further escalates the score by 4.96 through causal representation learning.
Though CMCIR~\citep{liu2023cross} applies the Swin-Transformer-L~\citep{liu2021swin} pretrained on ImageNet-22K dataset as the frame-level appearance extractor and employs the video Swin-B~\citep{liu2021videoswin} pretrained on Kinetics-600 as the clip-level motion feature extractor (more powerful than ours), \ourmeos with sample ResNet101~\citep{he2016deep} features still outperforms it with 2.64 in average. 
More analysis on TrafficQA and another evaluation on  Volleyball~\cite{ibrahim2016hierarchical} can be found in Appendix~\ref{sec: ap_traffic} and \ref{sec: ap_volleyball}, respectively.

\section{Conclusion}
In this paper, we have proposed to consider learning temporal causal representation under the non-invertible generation process, which is motivated by the common requirement of the temporal system, such as the visual perception process. 
We have established identifiability theories that allow for recovering the latent causal process with the nonlinear and non-invertible mixing function. Furthermore, based on this theorem, we proposed our approach, \caring, to leverage the temporal context to estimate the lost latent information. We have conducted a series of simulated experiments to verify the identifiability results of \ourmeos under the non-invertible generations and evaluated the learned representation in a complex and non-invertible traffic environment with real-world VideoQA tasks. 

\section*{Impact Statement} 

This study introduces both a theoretical framework and a practical approach for extracting causal representations from time-series data. 
Such advancements enable the development of more transparent and interpretative models, enhancing our grasp of causal dynamics in real-world settings. 
This approach may benefit many real-world applications, including healthcare, auto-driving, and finance, but it could also be used illegally.  For example, within the financial sphere, it can be harnessed to decipher ever-evolving market trends, optimizing predictions and thereby influencing investment and risk management decisions. 
However, it's imperative to note that any misjudgment of causal relationships could lead to detrimental consequences in these domains. Thus, establishing causal links must be executed with precision to prevent skewed or biased inferences.

Theoretically, though allowing for the non-invertible generation process, our theoretical assumptions still fall short of fully capturing the intricacies of real-world scenarios.  
For example, identifiability requires the absence of instantaneous causal relations, i.e., relying solely on time-delayed influences within the latent causal dynamics. Furthermore, we operate under the presumption that the number of variables remains consistent across different time steps, signifying that no agents enter or exit the environment.  
Moving forward, we aim to broaden our framework to ensure identifiability in more general settings, embracing instantaneous causal dynamics and the flexibility for variables to either enter or exit.

In our experiments, we evaluate our approach with both simulated and real-world datasets. However, our simulation relies predominantly on data points, creating a gap from real-world data. Concurrently, the real datasets lack the presence of ground truth latent variables. In the future,  we plan to develop a benchmark specifically tailored for the causal representation learning task. This benchmark will harness the capabilities of game engines and renderers to produce videos embedded with ground-truth latent variables.

\section*{Acknowledgments}
We would like to acknowledge the support from NSF Grant 2229881, the National Institutes of Health (NIH) under Contract R01HL159805, and grants from Apple Inc., KDDI Research Inc., Quris AI, and Florin Court Capital.

\bibliography{caring}
\bibliographystyle{icml2024}

\clearpage
\appendix

\onecolumn

\newcommand{\beginsupplement}{%
	\setcounter{table}{0}
	\renewcommand{\thetable}{A\arabic{table}}%

         \setcounter{equation}{0}
	\renewcommand{\theequation}{A\arabic{equation}}%
	
	\setcounter{figure}{0}
	\renewcommand{\thefigure}{A\arabic{figure}}%
	
	\setcounter{algorithm}{0}
	\renewcommand{\thealgorithm}{A\arabic{algorithm}}%
	
	\setcounter{section}{0}
	\renewcommand{\thesection}{A\arabic{section}}%

    \setcounter{theorem}{0}
    \renewcommand{\thetheorem}{A\arabic{theorem}}%

    \setcounter{corollary}{0}
    \renewcommand{\thecorollary}{A\arabic{corollary}}%

        \setcounter{lemma}{0}
    \renewcommand{\thelemma}{A\arabic{lemma}}%

}

\beginsupplement


\textit{\large Appendix for}\\ \ \\
      {\large \bf ``Learning Temporal Causal Representation under Non-Invertible Generation Process''}\

\vspace{.1cm}





\section{Identifiability Theory}
\label{ap:theory}
\subsection{Proof for Theorem 1}
\label{sec: app_th1}

{
Let us first shed light on the identifiability theory on the special case with $\tau=r+1$, i.e.,
\begin{equation}
    \label{eq:process_lag1_app}
   \mathbf{x}_t = \mathbf{g}(\mathbf{z}_{t:t-r}) , \quad z_{it} = f_i\left(\mathbf{z}_{t-1:t-r-1}, \epsilon_{it}  \right),\quad \mathbf{z}_t = \mathbf{m}(\mathbf{x}_{t:t-\mu}).
\end{equation}
}
 
{
\begin{theorem}[Identifiability under Non-invertible Generative Process] \label{Th1 in appendix} 
For a series of observations {$\mathbf{x}_t\in\mathbb{R}^d$} and estimated latent variables $\mathbf{\hat{z}}_{t}\in\mathbb{R}^n$ 
, suppose there exists function $\mathbf{\hat{g}},\mathbf{\hat{m}}$ which is subject to observational equivalence, 
\begin{equation}
    \label{eq:invert_lag1_app}
    \mathbf{x}_t = \mathbf{\hat{g}}(\mathbf{\hat{z}}_{t:t-r}),\quad \mathbf{\hat{z}}_{t} =\mathbf{\hat{m}}(\mathbf{x}_{t:t-\mu}).
\end{equation}
If assumptions
\begin{itemize}
    \item \{(Smooth and Positive Density) the probability density function of latent variables is third-order differentiable and positive in $R^n$,
    \item (conditional independence) the components of $\mathbf{\hat{z}}_{t}$ are mutually independent conditional on $\mathbf{\hat{z}}_{t-1:t-r-1}$,
    {\item (sufficiency) let $\eta_{kt} \triangleq \log p(z_{kt} | \mathbf{z}_{t-1:t-r-1})$, and
\begin{equation} \label{Eq:v_app}
\begin{aligned}
\mathbf{v}_{lt} 
\triangleq \Big(
\frac{\partial^2 \eta_{1t}}{\partial z_{1t} \partial z_{l,t-r-1}}, 
..., 
\frac{\partial^2 \eta_{nt}}{\partial z_{nt} \partial z_{l,t-r-1}} ,
\frac{\partial^3 \eta_{1t}}{\partial z_{1t}^2 \partial z_{l,t-r-1}}, 
..., 
\frac{\partial^3 \eta_{nt}}{\partial z_{nt}^2 \partial z_{l,t-r-1}} \Big)^\intercal,
\end{aligned}
\end{equation}
for $l=1,2,\cdots,n$. For each value of $\mathbf{z}_t$, there exists $2n$ different values of $z_{l,t-r-1}$ such that the $2n$ vector functions $\mathbf{v}_{lt}\in\mathbf{R}^{2n}$ are linearly independent,}
\end{itemize}
are satisfied, then $\mathbf{z}_{t}$ must be a component-wise transformation of a permuted version of $\mathbf{\hat{z}}_{t}$  with regard to context $\{\mathbf{x}_j\mid\forall j=t,t-1,\cdots,t-\mu-r\}$.
\end{theorem}
}

\begin{proof}
    For any $t$, combining Eq~\ref{eq:process_lag1_app} and Eq~\ref{eq:invert_lag1_app} gives 
    \begin{equation}
    \label{eq:xt_co1_app}
    \begin{aligned}
        \mathbf{z}_t  
        & = \mathbf{m}(\mathbf{x}_{t:t-\mu}) \\
        & = \mathbf{m}(\mathbf{\hat{g}}(\mathbf{\hat{z}}_t,\mathbf{\hat{z}}_{t-1:t-r}), \mathbf{x}_{t-1:t-\mu}) \\
        & = \mathbf{m}(\mathbf{\hat{g}}(\mathbf{\hat{z}}_t,\mathbf{\hat{m}}(\mathbf{x}_{t-1:t-\mu-1}), \cdots, \mathbf{\hat{m}}(\mathbf{x}_{t-r:t-\mu-r})), \mathbf{x}_{t-1:t-\mu}), \\
    \end{aligned}
    \end{equation}
    as well as $\mathbf{\hat{z}}_t = \mathbf{\hat{m}}(\mathbf{g}(\mathbf{z}_t,\mathbf{m}(\mathbf{x}_{t-1:t-\mu-1}), \cdots, \mathbf{m}(\mathbf{x}_{t-r:t-\mu-r})), \mathbf{x}_{t-1:t-\mu})$ similarly. Upon Eq~\ref{eq:xt_co1_app}, we have an unified partially invertible function $\mathbf{z}_t = \mathbf{h}(\mathbf{\hat{z}}_t|\mathbf{x}_{t-1:t-\mu-r})$ where $\mathbf{h}=\mathbf{m}\circ \hat{\mathbf{g}}$ with Jacobian $\frac{\partial\mathbf{z}_t}{\partial\mathbf{\hat{z}}_t} =\mathbf{H}_t(\mathbf{\hat{z}}_t;\mathbf{x}_{t-1:t-\mu-r})$. By \textit{partially invertible} it means that $\mathbf{z}$ and $\mathbf{\hat{z}}$ are in one-to-one correspondence for any context observations $\mathbf{x}_{t-1:t-\mu-r}$ that are fixed. One more thing to notify is that since $\mathbf{g},\mathbf{\hat{g}},\mathbf{m},\mathbf{\hat{m}}$ are second-order differentiable, the nested $\mathbf{h}$ is also second-order differentiable. Let us consider the mapping from joint distribution $(\mathbf{\hat{z}}_t,\mathbf{x}_{t-1:t-\mu-r-1})$ to $(\mathbf{z}_t,\mathbf{x}_{t-1:t-\mu-r-1})$, i.e.,
    \begin{equation}
    \label{eq: joint dist mix}
        P(\mathbf{z}_t,\mathbf{x}_{t-1:t-\mu-r-1})
        = P(\mathbf{\hat{z}}_t,\mathbf{x}_{t-1:t-\mu-r-1})
        \,/\, |\mathbf{J}_t|,
    \end{equation}
    where  
    \begin{equation}
        \mathbf{J}_t = \begin{bmatrix}
            \frac{\partial \mathbf{z}_t}{\partial \mathbf{\hat{z}}_t}&  \mathbf{0} \\
            * & \mathbf{I}
        \end{bmatrix},
    \end{equation}
    which is a lower triangle matrix, where $\mathbf{I}$ infers eye matrix and $*$ infers any possible matrix. Thus, we have determinant $|\mathbf{J}_t| = |\frac{\partial \mathbf{z}_t}{\partial \mathbf{\hat{z}}_t}| = |\mathbf{H}_t|$. Dividing both sides of Eq~\ref{eq: joint dist mix} by $P(\mathbf{x}_{t-1:t-\mu-r-1})$  gives
    \begin{equation}
    \label{eq:zt}
        \textbf{LHS} 
         = P( \mathbf{z}_t| \mathbf{x}_{t-1:t-\mu-r-1}) = P(\mathbf{z}_t|\mathbf{z}_{t-1:t-r-1}),
    \end{equation}
    since $\mathbf{z}_t $ and $ \mathbf{x}_{t-1:t-\mu-r-1} $ are independent conditioned on $ \mathbf{z}_{t-1:t-r-1}$. Similarly, $\textbf{RHS} = P( \mathbf{\hat{z}}_t| \mathbf{x}_{t-1:t-\mu-r-1}) = P(\mathbf{\hat{z}}_t|\mathbf{\hat{z}}_{t-r-1})$ holds true as well, which yields to
    \begin{equation}
        \label{eq:zt mix}
        P(\mathbf{z}_t|\mathbf{z}_{t-1:t-r-1}) = P(\mathbf{\hat{z}}_t|\mathbf{\hat{z}}_{t-1:t-r-1}) \,/\, |\mathbf{H}_t|.
    \end{equation}
    From a direct observation, if the components of $\hat{\mathbf{z}}_t$ are mutually independent given $\hat{\mathbf{z}}_{t-1:t-r-1}$, then for any distinct $i\neq j$, $\hat{z}_{it}$ and $\hat{z}_{jt}$ are conditionally independent given $(\hat{\mathbf{z}}_{t}\setminus \{\hat{z}_{it}, \hat{z}_{jt}\})\cup \hat{\mathbf{z}}_{t-1:t-r-1}$. This mutual independence of the components of $\hat{\mathbf{z}}_t$ based on $\hat{\mathbf{z}}_{t-1:t-r-1}$ implies two things:
    \begin{itemize}
    \item  $\hat{z}_{it}$ is independent from $\hat{\mathbf{z}}_{t}\setminus \{\hat{z}_{it}, \hat{z}_{jt}\}$ %
    conditional on $\hat{\mathbf{z}}_{t-1:t-r-1}$. Formally,
    $$p(\hat{z}_{it} \,|\, \hat{\mathbf{z}}_{t-1:t-r-1}) = p(\hat{z}_{it} \,|\, (\hat{\mathbf{z}}_{t}\setminus \{\hat{z}_{it}, \hat{z}_{jt}\})\cup \hat{\mathbf{z}}_{t-1:t-r-1}).$$
    \item $\hat{z}_{it}$ is independent from $\hat{\mathbf{z}}_{t}\setminus \{\hat{z}_{it}\}$ conditional on $\hat{\mathbf{z}}_{t-1:t-r-1}$. Represented as:
     $$p(\hat{z}_{it} \,|\, \hat{\mathbf{z}}_{t-1:t-r-1}) = p(\hat{z}_{it} \,|\, (\hat{\mathbf{z}}_{t}\setminus \{\hat{z}_{it}\})\cup \hat{\mathbf{z}}_{t-1:t-r-1}).$$
\end{itemize}
    From these two equations, we can derive:
    $$p(\hat{z}_{it} \,|\, (\hat{\mathbf{z}}_{t}\setminus \{\hat{z}_{it}\})\cup \hat{\mathbf{z}}_{t-1:t-r-1}) = p(\hat{z}_{it} \,|\, (\hat{\mathbf{z}}_{t}\setminus \{\hat{z}_{it}, \hat{z}_{jt}\})\cup \hat{\mathbf{z}}_{t-1:t-r-1}),$$
    which yields that $\hat{z}_{it}$ and $\hat{z}_{jt}$ are conditionally independent given $(\hat{\mathbf{z}}_{t}\setminus \{\hat{z}_{it}, \hat{z}_{jt}\})\cup \hat{\mathbf{z}}_{t-1:t-r-1}$ for $i\neq j$.
     Leveraging an inherent fact, i.e., if  $\hat{z}_{it}$ and $\hat{z}_{jt}$ are conditionally independent given $(\hat{\mathbf{z}}_{t}\setminus \{\hat{z}_{it}, \hat{z}_{jt}\})\cup \hat{\mathbf{z}}_{t-1:t-r-1}$, the subsequent equation arises:
     $$\frac{\partial^2 \log p(\hat{\mathbf{z}}_t, \hat{\mathbf{z}}_{t-1:t-r-1})}{\partial \hat{z}_{it} \partial \hat{z}_{jt}} = 0, $$ 
     assuming the cross second-order derivative exists. \\ \\
     Given that $p(\hat{\mathbf{z}}_t, \hat{\mathbf{z}}_{t-1:t-r-1}) = p(\hat{\mathbf{z}}_t \,|\, \hat{\mathbf{z}}_{t-1:t-r-1})p(\hat{\mathbf{z}}_{t-1:t-r-1})$ and $p(\hat{\mathbf{z}}_{t-1:t-r-1})$ remains independent of $\hat{z}_{it}$ or $\hat{z}_{jt}$, the above equality is equivalent to 
      \begin{equation} \label{Eq:iszero}
         \frac{\partial^2 \log p(\hat{\mathbf{z}}_t \,|\,\hat{\mathbf{z}}_{t-1:t-r-1})}{\partial \hat{z}_{it} \partial \hat{z}_{jt}} = 0.
     \end{equation}
     Referencing Eq~\ref{eq:zt mix},  it gets expressed as:
      \begin{equation}
     \log p(\hat{\mathbf{z}}_t \,|\, \hat{\mathbf{z}}_{t-1:t-r-1}) = \log p({\mathbf{z}}_t \,|\, {\mathbf{z}}_{t-1:t-r-1}) + \log |\mathbf{H}_t| = \sum_{k=1}^n \eta_{kt} + \log |\mathbf{H}_t|.
    \end{equation}
    The partial derivative w.r.t. $\hat{z}_{it}$ is presented below:
    \begin{flalign} \nonumber
      \frac{\partial \log p(\hat{\mathbf{z}}_t \,|\, \hat{\mathbf{z}}_{t-1:t-r-1})}{\partial \hat{z}_{it}} &=  \sum_{k=1}^n \frac{\partial \eta_{kt} }{\partial z_{kt}} \cdot \frac{\partial z_{kt}}{\partial \hat{z}_{it}} + \frac{\partial \log |\mathbf{H}_t|}{\partial \hat{z}_{it}} \\ \nonumber
      &= \sum_{k=1}^n \frac{\partial \eta_{kt}}{\partial z_{kt}} \cdot \mathbf{H}_{kit} + \frac{\partial \log |\mathbf{H}_t|}{\partial \hat{z}_{it}}.
     \end{flalign}
     The second-order cross derivative can be depicted as:
     \begin{flalign} \label{Eq:cross}
      \frac{\partial^2 \log p(\hat{\mathbf{z}}_t \,|\, \hat{\mathbf{z}}_{t-1:t-r-1})}{\partial \hat{z}_{it} \partial \hat{z}_{jt}}
      &= \sum_{k=1}^n \Big( \frac{\partial^2 \eta_{kt}}{\partial z_{kt}^2 } \cdot \mathbf{H}_{kit}\mathbf{H}_{kjt} + \frac{\partial \eta_{kt}}{\partial z_{kt}} \cdot \frac{\partial \mathbf{H}_{kit}}{\partial \hat{z}_{jt}} \Big) + \frac{\partial^2 \log |\mathbf{H}_t|}{\partial \hat{z}_{it} \partial \hat{z}_{jt}}.
     \end{flalign}
    According to Eq~\ref{Eq:iszero}, the right-hand side of the presented equation consistently equals 0. Therefore, for each index $l$ ranging from 1 to $n$, and every associated value of $z_{l,t-r-1}$, its partial derivative with respect to $z_{l,t-r-1}$ remains 0. That is,
     \begin{flalign}\label{eq:lind-ap}
      \sum_{k=1}^n \Big( \frac{\partial^3 \eta_{kt}}{\partial z_{kt}^2 \partial z_{l,t-r-1}} \cdot \mathbf{H}_{kit}\mathbf{H}_{kjt} + \frac{ \partial^2 \eta_{kt}}{\partial z_{kt} \partial z_{l,t-r-1}}  \cdot \frac{\partial \mathbf{H}_{kit}}{\partial \hat{z}_{jt} } \Big) \equiv 0,
     \end{flalign}
    where we leveraged the fact that entries of $\mathbf{H}_t$ do not depend on $z_{l,t-r-1}$. 
     Considering any given value of $\mathbf{z}_t$,
     {there exists at least $2n$ different values of $\mathbf{v}_{lt}$ such that they are linearly independent.}
     To make the above equation hold true, one has to set $\mathbf{H}_{kit}\mathbf{H}_{kjt} = 0$ or $i\neq j$. In other words, each row of $\mathbf{H}_t$ consists of at most a single non-zero entry, and $\mathbf{z}_{t}$ must be a component-wise transformation of a permuted version of $\mathbf{\hat{z}}_{t}$. 
\end{proof}

Note that in the proof of Theorem~\ref{Th1 in appendix}, we require the transition lag $\tau$ to be larger than the mixing lag $r=1$. When a mixing lag exists, the guarantee of identifiability requires dynamic information from a more previous time step. As long as this inequality $\tau>r$ is satisfied, the parameters $\tau$ can be extended to arbitrary numbers following a similar modification in Appendix~\ref{ap:moretau}. 

\subsection{Discussion for the sufficiency assumption.}

This assumption describes the changability of latent variables. Taking the video understanding as an example, the latent variables may represent the concepts. The linear independence of the latent variables means that there exists a characteristic of the concept that cannot be linearly represented by others. To further illustrate the sufficiency assumption, we give 2 examples \cite{yao2022temporally} to show when and when not the sufficiency assumption holds.

One possible distribution that breaks this assumption is the additive Gaussian noise. Denote $\mathbf{z}_ {h}$ as historical parents. Let $z_{kt} = q_k(\mathbf{z}_ {h})+\epsilon_{kt}$ where $\epsilon_ {kt}\sim N(0,1)$. In this case, we have $\eta_ {kt}=\log P(z_{kt}|\mathbf{z}_ {h}) = -\log\sqrt{2\pi} -\frac{(z_{kt}-q_k(\mathbf{z}_ {h}))^2}{2}$, and $\frac{\partial^2\log P(z_{kt}|\mathbf{z}_ {h})}{\partial^2 z_{kt}} =0$, which will violate the assumption.

On the opposite, if $\epsilon_{kt}$ subjects a zero mean generalized normal distribution: $P(\epsilon_{kt}) \propto e^{-\lambda|\epsilon_{kt}|^\beta}$ with $\lambda>0$ and $\beta>2$ and $\beta\not=3$. Let $z_{kt} = q_k(\mathbf{z}_ h)+\epsilon_{kt}$ in which $q$ is a linear function. If for each $z_{kt}$ there exists at least one $k'$ such that $c_{kk'}=\frac{\partial z_{kt}}{\partial z_{k',t-1}}\not=0$, the sufficiency assumption must hold.

In this case, we have 
\begin{equation}
    \frac{\partial^3 \eta_{kt}}{\partial^2 z_{kt}\partial z_{k',t-1}} = -\lambda \text{ sgn}(\epsilon_{kt})\beta(\beta-1)(\beta-2)|\epsilon_{kt}|^{\beta-3}c_{kk'}
\end{equation}
and 
\begin{equation}
    \frac{\partial^2 \eta_{kt}}{\partial z_{kt}\partial z_{k',t-1}} = -\lambda \beta(\beta-1)|\epsilon_{kt}|^{\beta-2}c_{kk'}. 
\end{equation}
We know that $|\epsilon_{lt}|^{\beta-2}$ and $|\epsilon_{lt}|^{\beta-2}$ are linearly independent since their ratio $|\epsilon_{lt}|$ is not constant. Besides, $|\epsilon_{lt}|^{\beta-2}$ and $|\epsilon_{lt}|^{\beta-2}$, with $l=1,2,\cdots,n$ are $2n$ linearly independent functions because of the different arguments involved. Suppose there exists $\alpha_{l1}, \alpha_{l2}$ for $l=1,2,\cdots,n$, such that the weighted sum with regard to $\mathbf{v}_{l,t}$ is zero. Thus, for any $k$ we have 
\begin{equation}
    \alpha_{k1}c_{kk'}|\epsilon_{kt}|^{\beta-2} + \alpha_{k2}c_{kk'}|\epsilon_{kt}|^{\beta-3} + \sum_{l\not=k}(\alpha_{l1}c_{lk'}|\epsilon_{lt}|^{\beta-2} + \alpha_{l2}c_{lk'}|\epsilon_{lt}|^{\beta-3}) =0.
\end{equation}

Since $|\epsilon_{kt}|^{\beta-2}$ and $|\epsilon_{kt}|^{\beta-3}$ with $l=1,2,\cdots,n$ are linearly independent and $c_{kk'}\not=0$, the make the above equation holds, we have $\alpha_{k1}=\alpha_{k2}=0$. As this applies to any k, we know that $\alpha_{l1}$ and $\alpha_{l2}$ must be 0, for all $l=1,2,\cdots,n$. That is, $\{\mathbf{v}_ {lt}\}$ is linearly independent. Thus, the sufficiency assumption holds.

Please note that the sufficiency assumption is crucial to the identifiability theory, yet not that restrictive. Even if it is not completely satisfied, we can still obtain some subspace identifiability~\cite{kong2022partial}.

\subsection{Discussion for the cross-time disentanglement}
This section demonstrates how the entanglements between variables across time steps are prevented. Generally speaking, if the information is lost in the transition from $\mathbf z_ {t}$ to $\mathbf x_ {t}$, we have to borrow the information from context such as $\mathbf z_ {t-1}$ to recover it. It is natural to receive information from $\mathbf z_ {t-1}$ in order to find the best estimator.

Specifically, let us consider a generating process $x_ {it} = g_i(\mathbf{z}_ {t}), z_ {it} = f_i(\mathbf{z}_ {t-1}, \epsilon_ {it})$, where $\mathbf{x}_ {t}\in R^d, \mathbf{z}_ {t}\in R^n$. Since $x_ {it}$ can be fully charactized by $\mathbf{z}_ {t}$, but not a function of $\mathbf{z}_ {t-1}$,  we have $\frac{\partial x_{i,t}}{\partial \mathbf{z}_ {t-1}}=0$. For the estimation process $x_ {i,t}=\hat{g}_ i(\mathbf{\hat{z}}_ {t})$, we have $\frac{\partial x_ {i,t}}{\partial \mathbf{z}_ {t-1}}=\frac{\partial x_ {i,t}}{\partial \mathbf{\hat{z}}_ {t}} \cdot\frac{\partial \mathbf{\hat{z}}_ {t}}{\partial \mathbf{z}_ {t-1}} = 0$ for $i=1,2,\cdots,d$. Formally, we have an equation as $\frac{\partial \mathbf{x}_ {t}}{\partial \mathbf{\hat{z}}_ {t}} \cdot \frac{\partial \mathbf{\hat{z}}_ {t}}{\partial \mathbf{z}_ {t-1}} = 0$, i.e., 
\begin{equation}
    \begin{bmatrix} \frac{\partial x_ {1,t}}{\partial \hat{z}_ {1t}} & \cdots & \frac{\partial x_ {1,t}}{\partial \hat{z}_ {nt}} \\ \vdots & \ddots & \vdots \\ \frac{\partial x_ {d,t}}{\partial \hat{z}_ {1t}} & \cdots & \frac{\partial x_ {d,t}}{\partial \hat{z}_ {nt}} \\ \end{bmatrix} \cdot \begin{bmatrix} \frac{\partial \hat{z}_ {1t}}{\partial z_ {1,t-1}} & \cdots & \frac{\partial \hat{z}_ {1t}}{\partial z_ {n,t-1}} \\ \vdots & \ddots & \vdots \\ \frac{\partial \hat{z}_ {nt}}{\partial z_ {1,t-1}} & \cdots & \frac{\partial \hat{z}_ {nt}}{\partial z_ {n,t-1}} \\ \end{bmatrix} = \mathbf{0}.
\end{equation}

If there exists at least $n$ different $i$ such that derivative of $x_ {it}$ with respect to $\hat{\mathbf{z}}_ t$ as $n$ vector functions $\begin{bmatrix} \frac{\partial x_ {i,t}}{\partial \hat{z}_ {1t}} & \cdots &\frac{\partial x_{i,t}}{\partial \hat{z}_ {nt}}\end{bmatrix}$ are linearly independent, we have 
\begin{equation}
    \begin{bmatrix} \frac{\partial \hat{z}_ {1t}}{\partial z_ {1,t-1}} & \cdots & \frac{\partial \hat{z}_ {1t}}{\partial z_ {n,t-1}} \\ \vdots & \ddots & \vdots \\ \frac{\partial \hat{z}_ {nt}}{\partial z_ {1,t-1}} & \cdots & \frac{\partial \hat{z}_ {nt}}{\partial z_ {n,t-1}} \\ \end{bmatrix} = \mathbf{0}
\end{equation}
holds true. If the rank $R(\frac{\partial \mathbf{x}_ {t}}{\partial \mathbf{\hat{z}}_ {t}})$ of the matrix of $\frac{\partial \mathbf{x}_ {t}}{\partial \mathbf{\hat{z}}_ {t}}$ is less than $n$, we have that the entanglement can only happened on the rest $n-R(\frac{\partial \mathbf{x}_ {t}}{\partial \mathbf{\hat{z}}_ {t}})$ dimensions.

Let us consider 2 extreme cases. First, if mixing function $\mathbf{g}$ is invertible, we have $R(\frac{\partial \mathbf{x}_ {t}}{\partial \mathbf{\hat{z}}_ {t}})=n$, and the entanglement between time steps are prevented. As another extreme case, in the NG setting as mentioned in Appendix~\ref{ap:synthetic}, we set one dimension of the latent variable that is totally lost during the mixing process. In this case, we have to use $\mathbf z_ {t-1}$ for its estimation. Even in this case, the unnecessary entanglements are prevented as well.

\subsection{Extension to Multiple Lags }
\label{ap:moretau}

{
\textbf{Multiple Transition Time Lag $\tau$.} For the sake of simplicity, we consider only one special case with $\tau=r+1$ in Theorem~\ref{Th1 in appendix}. Our identifiability theorem can be actually extended to arbitrary lags directly. For any given $\tau$, according to modularity, we have different conclusion at Eq~\ref{eq:zt} as
$\textbf{LHS} =P( \mathbf{z}_t|\mathbf{x}_{t-1:t-\mu-r-\tau}) = P(\mathbf{z}_t|\mathbf{z}_{t-1:t-r-\tau}). $
Similarity  $\textbf{RHS} = P( \mathbf{\hat{z}}_t|\mathbf{x}_{t-1:t-\mu-r-\tau}) = P(\mathbf{\hat{z}}_t|\mathbf{\hat{z}}_{t-1:t-r-\tau})$  holds true as well. In addition, some modifications are needed in sufficiency assumption, i.e., re-define $\eta_{kt} \triangleq \log p(z_{kt} | \mathbf{z}_{t-1:t-r-\tau})$ and there should be at least $2n$ linear independent vectors for $\mathbf{v}$ with regard to $z_{lt'}$ where $l=1,2,\cdots,n$ and $t-\tau \leq t'\leq t-r-1$. No extra changes are needed.
}

\textbf{Infinite Mixing Lag $r$.} Theorem \ref{Th1 in appendix} can also be easily extended to infinite mixing lag
since $\mathbf{\hat{z}}_t = \mathbf{h}(\mathbf{z}_t;\mathbf{x}_{<t})$ still exists when $r\rightarrow\infty$, and the theorem still holds true.

\subsection{Continuity for Permutation Invariance}
\label{ap: necessity of continuity}

Let us first give an extreme example to illustrate the importance of extra constraints for identifiability when the {probability density} of $\mathbf{z}_t$ is not non-zero everywhere in $\mathbb{R}^n$. Consider 4 independent random variables $u,v,x,y$ subject to standard normal distribution respectively. Suppose that there exists an invertible function $(x,y)=\mathbf{h}(u,v)$ satisfies
\begin{equation}
\label{eq: special case of indentifibility}
    \begin{cases}
        x = \mathbb{I}(x+y>0)\cdot u + \mathbb{I}(x+y\leq 0)\cdot v \\
        y = \mathbb{I}(x+y>0)\cdot v + \mathbb{I}(x+y\leq 0)\cdot u.
    \end{cases}
\end{equation}

Notice that the Jacobian from $(u,v)$ to $(x,y)$ contains at most one non-zero entry for each column or row. However, the result $(x,y)$ is still entangled, and the identifiability of $(u,v)$ is not achieved. What if now we notate latent variable  as $\mathbf{\hat{z}}=(u,v)$, estimated latent variable as $\mathbf{z}=(x,y)$ and the transition process with two mixing functions as $\mathbf{h}=\mathbf{g}^{-1}  \circ \mathbf{\hat{g}}$?

In the literature of nonlinear ICA, the gap between $\mathbf{H}_{ij}\cdot\mathbf{H}_{ik}=0$ when $j\not=k$ and identifiability is ill-discussed. In linear ICA, since the Jacobian is a constant matrix, these two statements are equivalent. Nevertheless, in nonlinear ICA, $\mathbf{H} = \frac{\partial\mathbf{z}}{\partial\mathbf{\hat{z}}}$ is not a constant, but a function of $\mathbf{\hat{z}}$, which may leads to the failure of identifiability as shown in Eq~\ref{eq: special case of indentifibility}.

The counterexamples can still be easily constructed even if function $\mathbf{h}$ is continuous. For brevity, let us denote a segment-wise linear indicator function as $f(u,v) = \min(\max(0, u+v+0.5),1)$, and we have $\mathbf{h}$ as
\begin{equation}
    \label{eq: special case of connected-path}
    \begin{cases}
        x = f(u,v)\cdot u + (1-f(u,v))\cdot v \\
        y = f(u,v)\cdot v + (1-f(u,v))\cdot u.
    \end{cases}
\end{equation}

When $u,v,x,y$ are independent uniform distributions on $[-2,-1]\cup[1,2]$, all conditions are still satisfied while the identifiability cannot be achieved.

{
To fill this gap, we provide two more assumptions. The domain $\mathcal{\hat{Z}}$ of $\mathbf{\hat{z}}$ should be path-connected, i.e., for any $\mathbf{\hat{z}}^{(1)}, \mathbf{\hat{z}}^{(2)} \in \mathcal{\hat{Z}}$, there exists a continuous path connecting $\mathbf{\hat{z}}^{(1)}$ and  $\mathbf{\hat{z}}^{(2)}$ with all points of the path in $\mathcal{\hat{Z}}$. In addition, the derivative of function $\mathbf{h}$ is not zero for any value of $\mathbf{\hat{z}} \in \mathcal{\hat{Z}}$ 
}

{
\begin{lemma}[Disentanglement with Continuity]
    \label{lemma: continuousity_app}
    For second-order differentiable invertible function $\mathbf{h}$ defined on a path-connected domain $\mathcal{\hat{Z}}\subseteq \mathbb{R}^n$ which satisfies $\mathbf{z} = \mathbf{h}(\mathbf{\hat{z}})$, suppose the non-degeneracy condition holds. If there exists at most one non-zero entry in each row of the Jacobian matrix $\mathbf{H}=\frac{\partial \mathbf{z}}{\partial \mathbf{\hat{z}}}$, the identifiability under Permutation Invariance can be established.
\end{lemma}
}

\begin{proof}
    {
    For any row $i$, $ \frac{\partial \mathbf{z}_i}{\partial \mathbf{\hat{z}}} = [ \frac{\partial \mathbf{z}_i}{\partial \mathbf{\hat{z}}_1}, \frac{\partial \mathbf{z}_i}{\partial \mathbf{\hat{z}}_2}, ..., \frac{\partial \mathbf{z}_i}{\partial \mathbf{\hat{z}}_n} ] \in \mathbb{R}^n$ is a n-dimensional variable. 
    Its image is a subspace as $\bigcup_{k=1}^{n} \left\{ (\frac{\partial \mathbf{z}_i}{\partial \mathbf{\hat{z}}_1}, \frac{\partial \mathbf{z}_i}{\partial \mathbf{\hat{z}}_2}, ..., \frac{\partial \mathbf{z}_i}{\partial \mathbf{\hat{z}}_n}) \in \mathbb{R}^n : \frac{\partial \mathbf{z}_i}{\partial \mathbf{\hat{z}}_j} = 0 \text{ for all } j \neq k, \text{ and } x_k \neq 0 \right\}$, since there exists at most one non-zero entry in each row of the Jacobian matrix $\mathbf{H}=\frac{\partial \mathbf{z}}{\partial \mathbf{\hat{z}}}$ and the derivative of function $\mathbf{h}$ is not zero for any value,  according to the non-degeneracy condition.
    }

    { We use proof by contradiction.
    Suppose there exist two different samples $\mathbf{a},\mathbf{b}\in\mathcal{Z}\subseteq \mathbb{R}^n$ with different non-zero entries $j\not=k$ subjects to
    }
    \begin{equation}
    \label{eq: contradict samples}
    {
            \left[\frac{\partial z_i}{\partial \hat{\mathbf{z}}}\bigg|_{\hat{\mathbf{z}}=\mathbf{a}}\right]_j  \not =0, \quad
            \left[\frac{\partial z_i}{\partial \hat{\mathbf{z}}}\bigg|_{\hat{\mathbf{z}}=\mathbf{b}}\right]_k  \not= 0  
        }
    \end{equation}
    {
    where $[\cdot]_j$ refers to the $j$-th entry of vector. Their values are respectively within $ \left\{ (0, 0,..., \frac{\partial {z}_i}{\partial {\hat{z}}_j}, 0 ,..., 0) \in \mathbb{R}^n : \frac{\partial {z}_i}{\partial {\hat{z}}_j} \neq 0 \right\}$ and $ \left\{ (0, 0,..., \frac{\partial {z}_i}{\partial {\hat{z}}_k}, 0 ,..., 0) \in \mathbb{R}^n : \frac{\partial {z}_i}{\partial {\hat{z}}_k} \neq 0 \right\}$.
    Clearly, there is no path from $\frac{\partial z_i}{\partial \hat{\mathbf{z}}}\big|_{\hat{\mathbf{z}}=\mathbf{a}} $ to $\frac{\partial z_i}{\partial \hat{\mathbf{z}}}\big|_{\hat{\mathbf{z}}=\mathbf{b}} $. 
    }
    {
    Since  $\mathbf{h}$ is a second-order differentiable invertible function, we have its derivative $\mathbf{h}'$ is also differentiable. 
    Thus, $\mathcal{\hat{Z}}\subseteq \mathbb{R}^n$ is a path-connected domain which denotes that the image of $ \frac{\partial {z}_i}{\partial \mathbf{\hat{z}}} $ is also path-connected.
    It will be violated that there is no path from $\frac{\partial z_i}{\partial \hat{\mathbf{z}}}\big|_{\hat{\mathbf{z}}=\mathbf{a}} $ to $\frac{\partial z_i}{\partial \hat{\mathbf{z}}}\big|_{\hat{\mathbf{z}}=\mathbf{b}} $ thus the proof is established.
    }
\end{proof}

{
When it comes to partially invertible function with regard to side information $\mathbf{c}$, the proof is the same with only a modification on conditions. That is, the path-connected domain assumption is applied to $(\mathbf{z}, \mathbf{c})$, and the infinite differentiability is extended to both $\mathbf{z}$ and $\mathbf{c}$, i.e., $\frac{\partial^2 z_i}{\partial a \partial b}$ for $a,b\in \{z|\mathbf{z}_i\}\times \{c|\mathbf{c}_i\}$ when $a\not=b$ exists.
}

{Let's further review the example we provided earlier. Examples in Eq~\ref{eq: special case of indentifibility} and Eq~\ref{eq: special case of connected-path} respectively demonstrate the scenarios where the assumptions of differentiability and connectivity fail, leading to the breakdown of identifiability. }

{
\begin{lemma}[Disentanglement with Continuity under Side Information]
    \label{lemma: coutinuous ++ app}
    For second-order differentiable invertible function $\mathbf{h}$ defined on a path-connected domain $\mathcal{\hat{Z}} \times \mathcal{C}\subseteq \mathbb{R}^{n+m}$ which satisfies $\mathbf{z} = \mathbf{h}(\mathbf{\hat{z}},\mathbf{c})$, suppose the non-degeneracy condition holds. If there exists at most one non-zero entry in each row of the Jacobian matrix $\mathbf{H}(\mathbf{c})=\frac{\partial \mathbf{z}}{\partial \mathbf{\hat{z}}}$, the identifiability under Permutation Invariance can be established.
\end{lemma}
}

\begin{proof}
{
    Suppose there exist two different samples $\mathbf{a},\mathbf{b}\in\mathcal{\hat{Z}} \times \mathcal{C}\subseteq \mathbb{R}^n$ with different non-zero entries $j\not=k$ subjects to
    \begin{equation}
    {
            \left[\frac{\partial z_i}{\partial (\hat{\mathbf{z}},\mathbf{c})}\bigg|_{(\hat{\mathbf{z}},\mathbf{c})=\mathbf{a}}\right]_j  \not =0, \quad
            \left[\frac{\partial z_i}{\partial (\hat{\mathbf{z}},\mathbf{c})}\bigg|_{(\hat{\mathbf{z}},\mathbf{c})=\mathbf{b}}\right]_k  \not= 0 .
        }
    \end{equation}
}

{
    Similar to Lemma~\ref{lemma: continuousity_app}, there exists no path between them because they are blocked in $\mathcal{\hat{Z}}$ alone. In the same way, since $\mathbf{h}$ is a second-order differentiable invertible function, and the non-degeneracy condition holds, the image of $\frac{\partial z_i}{\partial(\mathbf{\hat{z}},\mathbf{c})}$ is also path-connected. It will be violated and the proof is established.
}

\end{proof}

\subsection{Identifiability Benefits from  Non-Stationarity}
\label{sec: non-stationary}

We can further leverage the advantage of non-stationary data for identifiability. We rewrite $\mathbf{v}_{lt}$, which is defined in Eq~\ref{Eq:v_app}, as $\mathbf{s}_{lt}(u_r)$ in the $u_r$ context as
\begin{equation}\label{eq: s_lt}
\begin{aligned}
\mathbf{s}_{lt}(u_r)
\triangleq \Big(
\frac{\partial^2 \eta_{1t}(u_r)}{\partial z_{1t} \partial z_{l,t-r-1}}, 
..., 
\frac{\partial^2 \eta_{nt}(u_r)}{\partial z_{nt} \partial z_{l,t-r-1}} ,
\frac{\partial^3 \eta_{1t}(u_r)}{\partial z_{1t}^2 \partial z_{l,t-r-1}}, 
..., 
\frac{\partial^3 \eta_{nt}(u_r)}{\partial z_{nt}^2 \partial z_{l,t-r-1}} \Big)^\intercal.
\end{aligned}
\end{equation}

We also consider the version of subtraction $\mathring{\mathbf{s}}_{t}(u_r)$ from $u_r$ to $u_0$ without taking the derivative with respect to $z_{l,t-r-1}$ as
\begin{equation}\label{eq: ring s_t}
\begin{aligned}
\mathring{\mathbf{s}}_{t}(u_r)
\triangleq \Big(
\frac{\partial^2 \eta_{1t}(u_r)}{\partial z_{1t}}
-\frac{\partial^2 \eta_{1t}(u_0)}{\partial z_{1t}}, 
..., 
\frac{\partial^2 \eta_{nt}(u_r)}{\partial z_{nt}} 
-\frac{\partial^2 \eta_{nt}(u_0)}{\partial z_{nt}} , \\
\frac{\partial^3 \eta_{1t}(u_r)}{\partial z_{1t}^2} - 
\frac{\partial^3 \eta_{1t}(u_0)}{\partial z_{1t}^2}, 
..., 
\frac{\partial^3 \eta_{nt}(u_r)}{\partial z_{nt}^2} ^\intercal-
\frac{\partial^3 \eta_{nt}(u_0)}{\partial z_{nt}^2} \Big)^\intercal.
\end{aligned}
\end{equation}

As provided below, in our case, the identifiability of $\mathbf{z}_t$ is guaranteed by the linear independence of the whole function vectors $\mathbf{s}_{lt}(u_r)$ and $\mathring{\mathbf{s}}_{t}(u_r)$, with $l=1,2,...,n$ and every $u_r$. This linear independence is generally a much stronger condition. Theorem~\ref{Th1 in appendix} can be considered as a special case where the number of domains $u_r$ is $1$. In this case, only $\mathbf{s}_{lt}(u_0)$ in Eq~\ref{eq: s_lt} is utilized but $2n$ values of $z_{l,t-r-1}$ are required. Otherwise, in the nonstationary case, the domain information $u_r$ can increase the changeability of $s_{lt}(u_r)$. Besides, $\mathring{\mathbf{s}}_{t}(u_r)$ in Eq~\ref{eq: ring s_t} can also help to find more independent vectors to satisfy the sufficiency assumption.

\begin{corollary}[Identifiability under Non-Stationary Process]
Suppose $\mathbf{x}_t = \mathbf{g}(\mathbf{z}_{t:t-r})$, $\mathbf{z}_t = \mathbf{m}(\mathbf{x}_{t:t-\mu})$,  and that the conditional distribution $p(z_{kt} \,|\, \mathbf{z}_{t-1:t-r-1},\mathbf{u})$ may change across $a+1$ values of the auxiliary variable $\mathbf{u}$, denoted by $u_0$, $u_1$, ..., $u_a$. Suppose the components of $\mathbf{z}_t$ are mutually independent conditional on $\mathbf{z}_{t-1:t-r-1}$ with each auxiliary variable. Assume that the components of $\hat{\mathbf{z}}_t$ are also mutually independent conditional on $\hat{\mathbf{z}}_{t-1:t-r-1}$.  Suppose the domain is path-connected and $\mathbf{m},\mathbf{\hat{m}},\mathbf{g},\mathbf{\hat{g}}$ are second-order differentiable and their combination subjects to non-degenerate condition.
If there exists $2n$ different values of function vectors $\mathbf{s}_{lt}(u_r)$ or $\mathring{\mathbf{s}}_{t}(u_r)$ and $\mathring{\mathbf{s}}_{t}(u_r)$, with $l=1,2,...,n$ and every $u_r$, are linearly independent, then $\hat{\mathbf{z}}_t$ is a permuted invertible component-wise transformation of $\mathbf{z}_t$. 
\end{corollary}

\begin{proof}
    For any $t$ we have 
    \begin{equation}
    \begin{aligned}
        \mathbf{z}_t  
        & = \mathbf{m}(\mathbf{x}_{t:t-\mu}) \\
        & = \mathbf{m}(\mathbf{\hat{g}}(\mathbf{\hat{z}}_t,\mathbf{\hat{z}}_{t-1:t-r}), \mathbf{x}_{t-1:t-\mu}) \\
        & = \mathbf{m}(\mathbf{\hat{g}}(\mathbf{\hat{z}}_t,\mathbf{\hat{m}}(\mathbf{x}_{t-1:t-\mu-1}), \cdots, \mathbf{\hat{m}}(\mathbf{x}_{t-r:t-\mu-r})), \mathbf{x}_{t-1:t-\mu}), \\
    \end{aligned}
    \end{equation}
    as well as $\mathbf{\hat{z}}_t = \mathbf{\hat{m}}(\mathbf{g}(\mathbf{z}_t,\mathbf{m}(\mathbf{x}_{t-1:t-\mu-1}), \cdots, \mathbf{m}(\mathbf{x}_{t-r:t-\mu-r})), \mathbf{x}_{t-1:t-\mu})$ similarly. Thus, we have an unified partially invertible function $\mathbf{z}_t = \mathbf{h}(\mathbf{\hat{z}}_t|\mathbf{x}_{t-1:t-\mu-r})$ where $\mathbf{h}=\mathbf{m}\circ \hat{\mathbf{g}}$ with Jacobian $\frac{\partial\mathbf{z}_t}{\partial\mathbf{\hat{z}}_t} =\mathbf{H}_t(\mathbf{\hat{z}}_t;\mathbf{x}_{t-1:t-\mu-r})$. Let us consider the mapping from joint distribution $(\mathbf{\hat{z}}_t,\mathbf{x}_{t-1:t-\mu-r-1})$ to $(\mathbf{z}_t,\mathbf{x}_{t-1:t-\mu-r-1})$, i.e.,
    \begin{equation}
    \label{eq: joint dist mix ur}
        P(\mathbf{z}_t,\mathbf{x}_{t-1:t-\mu-r-1})
        = P(\mathbf{\hat{z}}_t,\mathbf{x}_{t-1:t-\mu-r-1})
        \,/\, |\mathbf{J}_t|,
    \end{equation}
    where  
    \begin{equation}
        \mathbf{J}_t = \begin{bmatrix}
            \frac{\partial \mathbf{z}_t}{\partial \mathbf{\hat{z}}_t}&  \mathbf{0} \\
            * & \mathbf{I}
        \end{bmatrix},
    \end{equation}
    which is a lower triangle matrix, where $\mathbf{I}$ infers eye matrix and $*$ infers any possible matrix. Thus, we have determinant $|\mathbf{J}_t| = |\frac{\partial \mathbf{z}_t}{\partial \mathbf{\hat{z}}_t}| = |\mathbf{H}_t|$. Dividing both sides of Eq~\ref{eq: joint dist mix ur} by $P(\mathbf{x}_{t-1:t-\mu-r-1},u_r)$  gives
    \begin{equation}
        \textbf{LHS} 
         = P( \mathbf{z}_t| \mathbf{x}_{t-1:t-\mu-r-1},u_r) = P(\mathbf{z}_t|\mathbf{z}_{t-1:t-r-1},u_r),
    \end{equation}
    since $\mathbf{z}_t $ and $ \mathbf{x}_{t-1:t-\mu-r-1} $ are independent conditioned on $ \mathbf{z}_{t-1:t-r-1}$ with any auxiliary variable $u_r$. Similarly, $\textbf{RHS} = P( \mathbf{\hat{z}}_t| \mathbf{x}_{t-1:t-\mu-r-1},u_r) = P(\mathbf{\hat{z}}_t|\mathbf{\hat{z}}_{t-r-1},u_r)$ holds true as well, which yields to
    \begin{equation}
        \label{eq:zt mix ur}
        P(\mathbf{z}_t|\mathbf{z}_{t-1:t-r-1},u_r) = P(\mathbf{\hat{z}}_t|\mathbf{\hat{z}}_{t-1:t-r-1},u_r) \,/\, |\mathbf{H}_t|.
    \end{equation}
    With conditional independence, we have
      \begin{equation}
         \frac{\partial^2 \log p(\hat{\mathbf{z}}_t \,|\,\hat{\mathbf{z}}_{t-1:t-r-1},u_r)}{\partial \hat{z}_{it} \partial \hat{z}_{jt}} = 0.
     \end{equation}
     Referencing Eq~\ref{eq:zt mix ur},  it gets expressed as:
      \begin{equation}
     \log p(\hat{\mathbf{z}}_t \,|\, \hat{\mathbf{z}}_{t-1:t-r-1},u_r) = \log p({\mathbf{z}}_t \,|\, {\mathbf{z}}_{t-1:t-r-1},u_r) + \log |\mathbf{H}_t| = \sum_{k=1}^n \eta_{kt}(u_r) + \log |\mathbf{H}_t|.
    \end{equation}
    
    The second-order derivative is
     \begin{equation}
     \label{eq: cross2 ur}
      \sum_{k=1}^n \Big( \frac{\partial^2 \eta_{kt}(u_r)}{\partial z_{kt}^2 } \cdot \mathbf{H}_{kit}\mathbf{H}_{kjt} + \frac{\partial \eta_{kt}(u_r)}{\partial z_{kt}} \cdot \frac{\partial \mathbf{H}_{kit}}{\partial \hat{z}_{jt}} \Big) + \frac{\partial^2 \log |\mathbf{H}_t|}{\partial \hat{z}_{it} \partial \hat{z}_{jt}}\equiv 0.
     \end{equation}
        
    The right-hand side of the presented equation consistently equals 0. Therefore, for each index $l$ ranging from 1 to $n$, and every associated value of $z_{l,t-r-1}$, its partial derivative with respect to $z_{l,t-r-1}$ remains 0. That is,
     \begin{equation}
     \label{eq: cross3 ur}
      \sum_{k=1}^n \Big( \frac{\partial^3 \eta_{kt}(u_r)}{\partial z_{kt}^2 \partial z_{l,t-r-1}} \cdot \mathbf{H}_{kit}\mathbf{H}_{kjt} + \frac{ \partial^2 \eta_{kt}(u_r)}{\partial z_{kt} \partial z_{l,t-r-1}}  \cdot \frac{\partial \mathbf{H}_{kit}}{\partial \hat{z}_{jt} } \Big) \equiv 0,
     \end{equation}
     where we leveraged the fact that entries of $\mathbf{H}_t$ do not depend on $z_{l,t-r-1}$.

     Again start from Eq~\ref{eq: cross2 ur}. Using the fact that $\mathbf{H}_t$ is not affected by the auxiliary variable, we can subtract the equation with $u_0$ from that of $u_r$. We have 

    \begin{equation}
    \label{eq: cross2 ur-u0}
      0 = 
      \sum_{k=1}^n \Big( 
      \Big(
      \frac{\partial^2 \eta_{kt}(u_r)}{\partial z_{kt}^2 }-
      \frac{\partial^2 \eta_{kt}(u_0)}{\partial z_{kt}^2 }
      \Big)
      \cdot \mathbf{H}_{kit}\mathbf{H}_{kjt} + 
      \Big(
      \frac{\partial \eta_{kt}(u_r)}{\partial z_{kt}}-
      \frac{\partial \eta_{kt}(u_0)}{\partial z_{kt}}
      \Big)
      \cdot \frac{\partial \mathbf{H}_{kit}}{\partial \hat{z}_{jt}} \Big) .
     \end{equation}

     Considering any given value of $\mathbf{z}_t$,
     there exists at least $2n$ different values of $\mathbf{s}_{lt}$ or $\mathring{\mathbf{s}}_{t}$, which corresponds to Eq~\ref{eq: cross3 ur} and Eq~\ref{eq: cross2 ur-u0} respectively,  such that they are linearly independent.
     To make the above equation hold true, one has to set $\mathbf{H}_{kit}\mathbf{H}_{kjt} = 0$ or $i\neq j$. In other words, each row of $\mathbf{H}_t$ consists of at most a single non-zero entry, and $\mathbf{z}_{t}$ must be a component-wise transformation of a permuted version of $\mathbf{\hat{z}}_{t}$. 
\end{proof}

\section{Synthetic experiments}

\subsection{Synthetic Dataset Generation}\label{ap:synthetic}

In this section, we give 2 representative simulation settings for \textbf{NG} and \textbf{NG-TDMP} respectively to reveal the identifiability results. For each synthetic dataset, we set latent space to be $3$, i.e., $\mathbf{x}_t \in \mathcal{X} \subseteq \mathbb{R}^3$.

\paragraph{Non-invertible Generation}

For \textbf{NG}, we set the transition lag as $\tau=1$. We first generate $10,000$ data points from a uniform distribution as the initial state $\mathbf{z}_0\sim U(0,1)$. For $t=1,\cdots,9$, each latent variable $\mathbf{z}_t$ will be generated from the proceeding latent variable $\mathbf{z}_{t-1}$ through a nonlinear function $\mathbf{f}$ with a non-additive zero-biased Gaussian noise $\mathbf{\epsilon}_t$ ($\sigma=0.1$), i.e., $\mathbf{z}_t = \mathbf{f}(\mathbf{z}_t, \epsilon_t)$. To introduce the non-invertibility, the mixing function $\mathbf{g}$ leverages only the first two entries of the latent variables to generate the 2-d observation $\mathbf{z}_t = \mathbf{g}(x_{1,t},x_{2,t}) \in \mathcal{Z} \subseteq \mathbb{R}^2$. 

\paragraph{Time-Delayed Mixing Process}

For \textbf{UG-TDMP}, we set the transition lag as $\tau=1$ and mixing lag $r=2$. Similar to the Non-invertible Generation scenario, we generate the initial states from a uniform distribution and the subsequent latent variables following a nonlinear transition function. The noise is also introduced in a nonlinear Gaussian ($\sigma=0.1$) way. The mixing process is a nonlinear function with regard to $\mathbf{z}_t$ plus a side information from previous steps $\mathbf{z}_{t-1:t-2}$, i.e., \begin{equation}
    \mathbf{x}_t = A_{3\times 3}\cdot\sigma\big(B_{3\times 3}\cdot\sigma(C_{3\times 3}\cdot \mathbf{z}_t)\big) + \begin{bmatrix}
    0 \\ 0 \\ D_{3\times 1}\mathbf{z}_{t-1} + E_{3\times 1}\mathbf{z}_{t-2}
    \end{bmatrix},
\end{equation}
where $\sigma$ refers to the ReLU function and the capital characters refer to matrices. Note that we make two modifications to show the advantage of \caring. The reason we consider larger mixing lag is that it is a much more difficult scenario to handle, with more distribution from the mixing process and less dynamic information from transition. We run experiments in both scenarios with different transition and mixing lag. Besides, we also find out that even without time-lagged latent variables in the decoder, it leads to a smaller model that is more stable and easy to train. Refer to Table~\ref{tab: ablation tdmp} for a detailed ablation study.

\begin{table}[t]
    \centering
    \begin{tabular}{c|c|c}
        \toprule
        \textbf{setting}  &  $\tau=1,r=2$ & $\tau=2,r=1$ \\
        \midrule
        \ourmeos &0.9436 & 0.9131 \\
        \ourmeos(lagged decoder) &0.9250 & 0.9220\\
        TDRL &0.8947 & 0.7519\\
        \bottomrule
    \end{tabular}
    \caption{Ablation study on different settings for \textbf{UG-TDMP}. (a) The second column is a more difficult scenario compared to the first, where the performance of \ourmeos remains good while that of baseline decreases significantly. (b) Omit the time-lagged latent variables in the decoder will not damage the performance much, but one can enjoy the benefits from a much simpler model.}
    \label{tab: ablation tdmp}
\end{table}

\paragraph{Post-processing Precedure}
During the generating process, we did not explicitly enforce the data to meet the constraint $\mathbf{z}_t=\mathbf{m}(\mathbf{x}_{t:t-\mu})$. On the contrary, we implement a checker to filter the data that is qualified. To be more precise, we do linear regression from $\mathbf{x}_{t:t-\mu}$ to $\mathbf{z}_t$ to figure out how much information of latent variables can be recovered from observation series in the best case. We choose the smallest $\mu$ when the amount of information that can be recovered is acceptable. We set $\mu=2$ for \textbf{UG} and $\mu=4$ for \textbf{UG-TDMP}.


\subsection{Implementation Details}
\paragraph{Network Architecture} To implement the Sequence-to-Step encoder, we leverage the \textit{torch.unfold} to generate the nesting observations. Let us denote $\mathbf{x}_t^{(\mu)} = [\mathbf{x}_t,\cdots,\mathbf{x}_{t-\mu}]$ as inputs. For the time steps that do not exist, we simply pad them with zero. Refer to Table~\ref{tab:arch-details 1} for detailed network architecture.

\paragraph{Training Details} The models were implemented in PyTorch 1.11.0. An AdamW optimizer is used for training this network. We set the learning rate as $0.001$ and the mini-batch size as $64$. We train each model under four random seeds ($770, 771, 772, 773$) and report the overall performance with mean and standard deviation across different random seeds.

\begin{table}[ht]
\caption{ Architecture details. BS: batch size, T: length of time series, i\_dim: input dimension, o\_dim: output dimension, z\_dim: latent dimension, LeakyReLU: Leaky Rectified Linear Unit.}
\label{tab:arch-details 1}
\begin{tabularx}{\textwidth}{@{\extracolsep{\fill}}lll}
\toprule
\textbf{Configuration} & \textbf{Description} &  \textbf{Output} \\
\toprule
\toprule
\textbf{1. Sequence-to-Step Encoder} &  Encoder for Synthetic Data & \\
\toprule
Input: $\mathbf{x}^{(\mu)}_{1:T}$ & Observed time series & BS $\times$ T $\times$ i\_dim \\
Dense & 128 neurons, LeakyReLU & BS $\times$ T $\times$ 128\\
Dense & 128 neurons, LeakyReLU & BS $\times$ T $\times$ 128 \\
Dense & 128 neurons, LeakyReLU & BS $\times$ T $\times$ 128 \\
Dense & Temporal embeddings & BS $\times$ T $\times$ z\_dim \\
\toprule
\toprule
\textbf{2. Step-to-Step Decoder} & Decoder for Synthetic Data & \\
\toprule
Input: $\hat{\mathbf{z}}_{1:T}$ & Sampled latent variables & BS $\times$ T $\times$ z\_dim \\
Dense & 128 neurons, LeakyReLU & BS $\times$ T $\times$ 128 \\
Dense & 128 neurons, LeakyReLU & BS $\times$ T $\times$ 128 \\
Dense & i\_dim neurons, reconstructed $\mathbf{\hat{x}}_{1:T}$ & BS $\times$ T $\times$ o\_dim \\
\toprule
\toprule
\textbf{3. Factorized Inference Network} & Bidirectional Inference Network & \\
\toprule
Input & Sequential embeddings & BS $\times$ T $\times$ z\_dim \\
Bottleneck & Compute mean and variance of posterior & $\mathbf{\mu}_{1:T}, \mathbf{\sigma}_{1:T}$ \\
Reparameterization & Sequential sampling & $\hat{\mathbf{z}}_{1:T}$ \\
\toprule
\toprule
\textbf{4. Modular Prior} & Nonlinear Transition Prior Network & \\
\toprule
Input & Sampled latent variable sequence $\hat{\mathbf{z}}_{1:T}$ & BS $\times$ T $\times$ z\_dim \\
InverseTransition & Compute estimated residuals $\hat{\epsilon}_{it}$ & BS $\times$ T $\times$ z\_dim \\
JacobianCompute & Compute $\log \left(\lvert \det \left(\mathbf{J}\right) \rvert \right)$ & BS\\
\bottomrule
\end{tabularx}
\end{table}

\begin{table}[ht]\renewcommand\tabcolsep{7.0pt}\renewcommand\arraystretch{0.95}
\begin{center}
\caption{\textbf{MCC scores of synthetic datasets with higher dimension.} }
\vspace{-0.1cm}
\label{tab: higher dimension} 
\begin{tabular}{c|c|c}
\toprule
Dimension & \ourmeos & TDRL \\
\midrule
6	& 0.9199	& 0.6329 \\
12	& 0.9366	& 0.6155 \\
18	& 0.7175	& 0.5265 \\
\bottomrule
\end{tabular}
\end{center}
\vspace{-0.2cm}
\end{table}

\subsection{{Exploration on higher dimension.}}
To demonstrate the scalability of our method, we have included experiments with higher dimensions. We keep the experimental setup consistent with NG and set the dimensions of latent variables to be $6, 12, 18$ and that of observation to be $4, 8, 12$, respectively. The transition function is a permutation function with a shift of $2, 4, 6$ dimensions respectively. As shown in Table~\ref{tab: higher dimension}, CaRiNG can achieve a consistent improvement over the baseline TDRL when using various dimensions. When the dimension is too high, although the performance of both CaRiNG and TDRL drops because of the complexity of the model, we demonstrate that CaRiNG still benefits from contextual information. This indicates that CaRiNG is scalable and robust to the dimensionality of the latent variables.

\subsection{Model Selection with Varying $\mu$}
\label{sec: mu_app}

{In this subsection, we will discuss the preliminary experiment that was instrumental in the model selection process for our application in the NG-TDMP settings. The experiment focused on evaluating the performance of the model with varying lengths of time lag $\mu$. }

{
Our findings indicate that an increase in $\mu$ does not always correlate with enhanced model performance. We observed that the effectiveness of each latent variable diminishes as the time lag $\mu$ increases. In practical applications, this motivates a strategy of model selection where an appropriate value of $\mu$ is chosen based on the model's performance. The following table summarizes our experimental results:
}

\begin{table}[h]
\centering
\begin{tabular}{c|c|c|c}
\toprule
$\mu$ & 3 & 4 & 5 \\
\midrule
Accuracy (\%) & 0.88 & 0.92 & 0.92 \\
\bottomrule
\end{tabular}
\caption{{Impact of varying $\mu$ on model performance in NG-TDMP settings.}}
\label{tab:mu_experiment}
\end{table}

These results suggest that while a larger $\mu$ might imply a more extensive recovery of context information, it can also introduce inefficiencies in information recovery, potentially adding noise and impeding model training.

\section{Real-world Experiments on TrafficQA}
\label{sec: ap_traffic}

\subsection{Implementation Details}

We choose HCRN~\citep{le2020hierarchical} (without classification head) as the encoder backbone of $\ourmeos$ on the real-world dataset: SUTD-TrafficQA. Given that HCRN is an encoder that calculates the cross attention between visual input and text input sequentially, we apply a decoder, which shares the same structure as the Step-to-Step Decoder shown in Table~\ref{tab:arch-details 1} to reconstruct the visual feature embedded with the temporal information. As it goes to transition prior, we use the Modular Prior shown in Table~\ref{tab:arch-details 1}. This encoder-decoder structure can guide the model to learn the hidden representation with identifiable guarantees under the non-invertible generation process.

\subsection{More Qualitative Results}
\label{sec:visualization}

\begin{figure*}[t]
\begin{center}
\centerline{\includegraphics[width=1.0\linewidth]{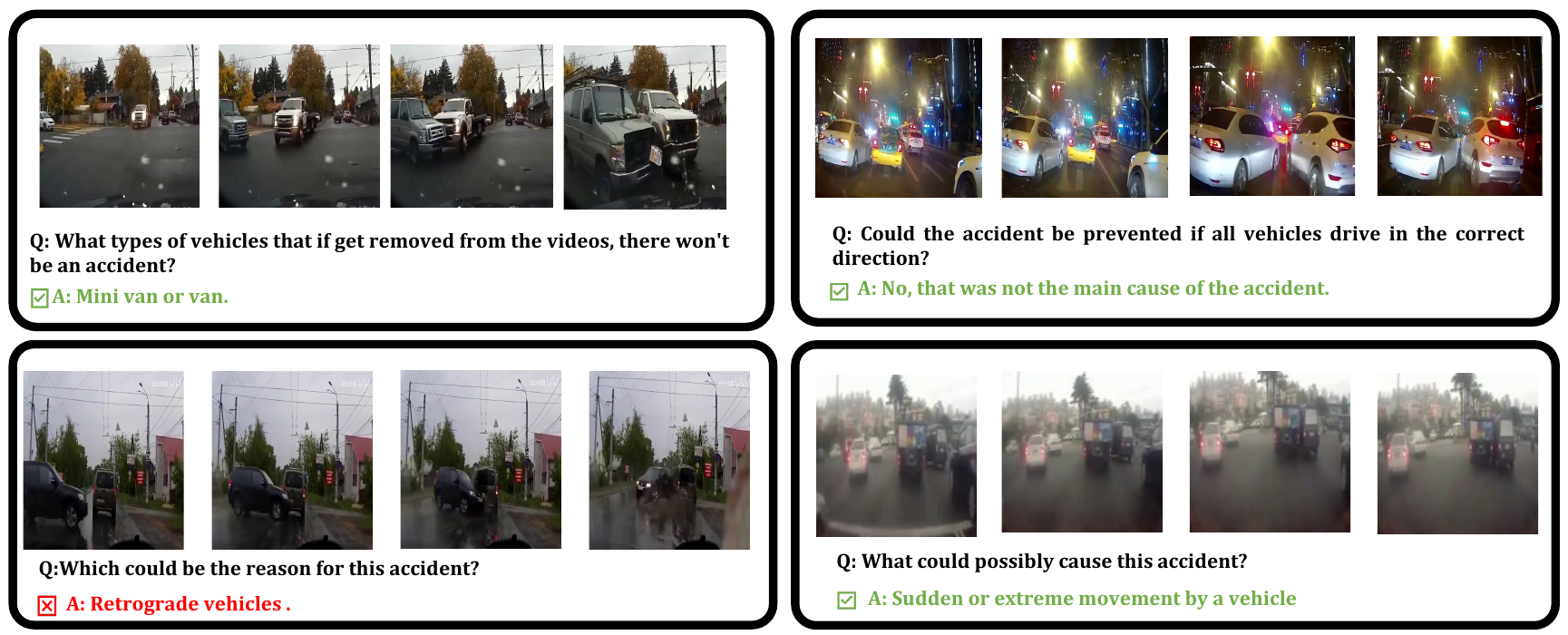}}
\caption{\textbf{Qualitative results on SUTD-TrafficQA dataset.}  We provide some positive examples and also fail cases to analyze our model.}
\label{fig:taffic_vis}
\end{center}
\end{figure*}

As shown in Figure~\ref{fig:taffic_vis}, we provide some positive examples and also fail cases to analyze our model. From the top two examples, we can find that our method can solve the occlusions well.  From the bottom right one, we find that our model can solve the blurred situation. However, when the alignment between visual and textual domains is difficult. The model may fail. 

\subsection{Computation Cost Comparison}
\label{sec: cost_app}

{
We provide the comparisons between the computational cost of the \ourmeos model and HCRN to analyze our efficiency. 
As shown in Table~\ref{table:comparison_cost}, we provide a detailed comparison of the number of parameters, training time, and inference efficiency. It is important to note that while the \ourmeos model requires a longer training time due to the application of normalizing flow for calculating the Jacobian matrix, its inference efficiency remains on par with HCRN, as the normalizing flow is utilized only for calculating KL loss and not during inference.}

\begin{table}[h]
\centering
\begin{tabular}{l|cc}
    \toprule
    \textbf{Method} & \textbf{HCRN} & \textbf{CaRiNG} \\
    \midrule
    Number of Parameters & 42,278,786 & 43,721,954 \\
    Training Time per Epoch & 6min 54s/epoch & 13min 26s/epoch \\
    Inference Time per Epoch & 49s/epoch & 49s/epoch \\
    \bottomrule
\end{tabular}
\caption{{Comparative Analysis of HCRN and \caring\ Models}}
\label{table:comparison_cost}
\end{table}

{This analysis clearly demonstrates that the increased training time for the \ourmeos model is offset by its comparable inference efficiency, highlighting its practical applicability in scenarios where inference time is critical.}

{
\subsection{Evaluation of Identifiability in the QA Benchmark}
\label{sec: idenqa_app}
}

{In the context of real-world applications, particularly in scenarios lacking ground truth for rigorous metrics like MCC, alternative evaluation strategies become essential. we leverage proxy metrics to assess the performance of the proposed algorithm, focusing on two pivotal aspects: disentanglement and reconstruction ability of the learned representations. 
Intuitively, as delineated in Theorem~\ref{Th1 in appendix} and detailed in Section~\ref{sec:method}, a representation can be considered identifiable if it possesses the dual capability of fully reconstructing the observation while also achieving disentanglement.
Thus, as a supplement to the accuracy we used before, we benchmark disentanglement and reconstruction ability as side evidence to support that the improvement is caused by better identifiability. 
}

\begin{figure}[h]
\centering
\includegraphics[width=0.5\textwidth]{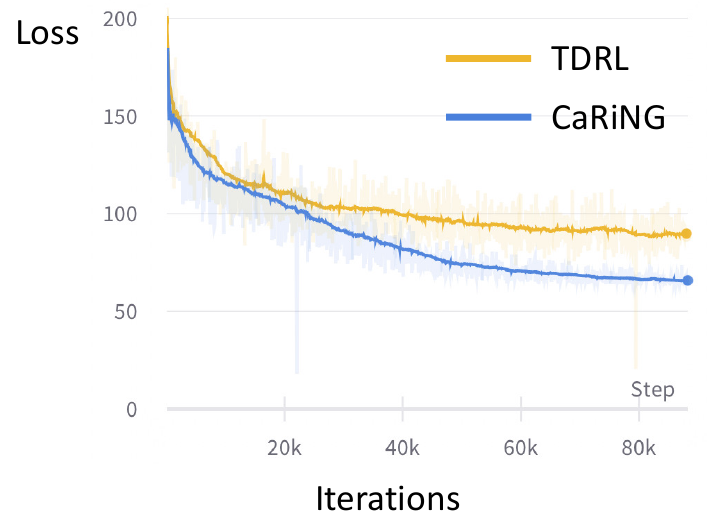}
\caption{{Comparative analysis of disentanglement and reconstruction abilities of different methods.}}
\label{fig:performance_comparison_loss}
\end{figure}

{
We use the ELBO loss as a proxy metric to evaluate the identifiability. Figure \ref{fig:performance_comparison_loss} illustrates our method's performance compared to the baseline TDRL method. The results clearly show that our approach exhibits superior disentanglement and reconstruction abilities. This evidence suggests that the advantage of our proposed algorithm is primarily attributed to its enhanced identifiability and effective disentanglement of data representations.
}

{
\subsection{Parameter analysis on $\tau$ }
\label{sec: tau_app}
}

{In this section, we present the results of our parameter analysis conducted on the SUTD-TrafficQA dataset, focusing on the impact of varying the time lag $\tau$. The study aimed to assess the robustness of our model to changes in the time lag parameter. As the table below illustrates, the model demonstrates consistent accuracy across different values of $\tau$, indicating robustness to the variation in time lag.}

\begin{table}[t]
\centering
\begin{tabular}{c|c|c|c}
\toprule
$\tau$ & 1 & 2 & 3 \\
\midrule
Accuracy (\%) & 41.22 & 41.23 & 41.27 \\
\bottomrule
\end{tabular}
\caption{{Parameter analysis results of $\tau$ on model accuracy in the SUTD-TrafficQA dataset.}}
\label{tab:ablation_tau}
\end{table}

\section{Real-world Experiments on the Volleyball Dataset}
\label{sec: ap_volleyball}

\subsection{Dataset}

The volleyball dateset~\citep{ibrahim2016hierarchical} is a video action recognition dataset with 4,830 clips from 55 videos. There are 8 group activity labels, including 4 main activities (set, spike, pass, win-point) that are divided into two subgroups, left and right. Two formats for inputs are provided: RGB videos and keypoints time series. In our setting, we simply use key points as the input. We utilized the 'original' split of the Volleyball dataset in which all videos were randomly assigned, consisting of 39 training videos and 16 testing videos. We adopt this dataset due to the complex occlusion in the sports which is aligned with our non-invertible generation setting. 

\subsection{Implementation Details}
The method is implemented using a VAE network. Specifically, the Sequence-to-Step Encoder processes the data by first flattening the features from all time steps. Then, following ~\citep{zhou2022composer}, we apply a Composer to incorporate the interactions with fine-grained information. Subsequently, we aggregate the contextual information through an MLP, mapping from a space of $\mathbb{R}^{T\times d}$ to $\mathbb{R}^{d}$.  
The Step-to-Step Decoder is also an MLP network mapping from $\mathbb{R}^{d}$ to $\mathbb{R}^{d}$. 
We adopt the same Modular Prior network as Table~\ref{tab:arch-details 1}. 
For the implementation of TDRL, the only difference is the removal of temporal dependencies during the encoding process of the model (don't aggregate the contextual information).

\subsection{Results and Analysis}

As shown in Table~\ref{tab: volleyball},  we observe that \ourmeos achieve consistent performance improvement on both person and group activity accuracy. It indicates that the temporal context is useful in the temporal dynamic modeling.  Though the goal of this task is the group activity recognition, we found that the person activity accuracy achieves more improvement. It is not surprising since our method ensures better disentanglement and identification of the latent variables of the group activity, i.e., containing the information of persons. 

\begin{table}[h]
\centering
\begin{tabular}{c|c|c}
\toprule
Method & \ourmeos & TDRL  \\
\midrule
Group Activity Top1 Accuracy(\%) & 93.044 & 92.895  \\
Group Activity Top3 Accuracy(\%) & 99.028 & 98.280  \\
Person Activity Top1 Accuracy(\%) & 74.551 & 73.286  \\
Person Activity Top3 Accuracy(\%) & 98.087 & 96.634  \\
\bottomrule
\end{tabular}
\caption{{Model accuracy in the Volleyball dataset.}}
\label{tab: volleyball}
\end{table}

\section{Related Work}

\subsection{Causal Discovery with Latent Variables}
Some studies have aimed to discover causally related latent variables, such as \cite{silva2006learning,kummerfeld2016causal,huang2022latent} leverage the vanishing Tetrad conditions~\cite{spearman1928pearson} or rank constraints to identify latent variables in linear-Gaussian models, and \cite{shimizu2009estimation,cai2019triad,xie2020generalized,xie2022identification} draw upon non-Gaussianity in their analysis for linear, non-Gaussian scenarios.
Furthermore, some methods aim to find the structure beyond the latent variables, resulting in the hierarchical structure. 
Some hierarchical model-based approaches assume tree-like configurations, such as \cite{pearl1988probabilistic,zhang2004hierarchical,choi2011learning,drton2017marginal}, while the other methods assume a broader hierarchical structure \cite{xie2022identification,huang2022latent}.
However, these methods remain confined to linear frameworks and face escalating challenges with intricate datasets, such as videos.

\subsection{Nonlinear ICA for Time Series Data}
Nonlinear ICA represents an alternative methodology to identify latent causal variables within time series data. Such methods leverage auxiliary data—like class labels and domain indices—and impose independence constraints to facilitate the identifiability of latent variables. To illustrate:
Time-contrastive learning (TCL \citep{hyvarinen2016unsupervised}) adopts the independent sources premise and capitalizes on the variability in variance across different data segments.  Furthermore, Permutation-based contrastive (PCL \citep{hyvarinen2017nonlinear}) puts forth a learning paradigm that distinguishes genuine independent sources from their permuted counterparts. Furthermore, 
i-VAE \citep{khemakhem2020variational} utilizes deep neural networks, VAEs, to closely approximate the joint distribution encompassing observed and auxiliary non-stationary regimes. Recent work, exemplified by LEAP~\citep{yao2021learning}, has tackled both stationary and non-stationary scenarios in tandem. In the stationary context, LEAP postulates a linear non-Gaussian generative process. For the non-stationary context, it assumes a nonlinear generative process, gaining leverage from auxiliary variables. Advancing beyond LEAP, TDRL \citep{yao2022temporally} initially extends the linear non-Gaussian generative assumption to a nonlinear formulation for stationary scenarios. Subsequently, it broadens the non-stationary framework to accommodate structural shifts, global alterations, and combinations thereof. Additionally, CITRIS \citep{lippe2022citris,lippe2022icitris} champions the use of intervention target data to precisely identify scalar and multi-dimensional latent causal factors. However, a common thread across these methodologies is the presumption of an invertible generative process, a stance that often deviates from the realities of actual data.
{Besides, \citep{hartford2022beyond} demonstrates that under a non-invertible scenario without extra information, identifiability can be only achieved in a subspace where bijective mapping exists. Their work provides additional support for the importance of addressing non-invertibility.}

\subsection{Temporal modeling}

Sequential Variational Autoencoders have gained significant popularity for their applications in temporal modeling, including generation, representation, and prediction. Variational RNN~\citep{sequential_vae} introduces the Variational Autoencoders into Recurrent Neural Networks, enabling variational inference on time series data. SRNN~\citep{fraccaro2016sequential} further utilizes the concept of SSM (State Space Model) for temporal modeling. In addition, SKD~\cite{berman2022multifactor} utilizes a structured Koopman autoencoder to achieve multifactor sequential disentanglement. However, none of these methods incorporates a transition function for capturing the temporal dynamics of multivariate data. By integrating a transition function with independent noise through normalizing flow~\cite{rezende2015variational,ziegler2019latent}, our model can effectively track and represent the causal relations of latent variables over time. Such enhancement positions \caring ~~as a method focused on learning causal representations with clear identifiability guarantees, marking a departure from the generation-centric objectives commonly seen in traditional VAE-based methods.

\end{document}

%% file: math_commands.tex
\usepackage{amsmath,amsfonts,bm}

\def\eqref#1{equation~\ref{#1}}

\def\1{\bm{1}}

\DeclareMathAlphabet{\mathsfit}{\encodingdefault}{\sfdefault}{m}{sl}
\SetMathAlphabet{\mathsfit}{bold}{\encodingdefault}{\sfdefault}{bx}{n}

